\pdfoutput=1

\documentclass[11pt]{article}

\usepackage[preprint]{acl}
\usepackage{times}
\usepackage{latexsym}

\usepackage[T1]{fontenc}

\usepackage[utf8]{inputenc}

\usepackage{microtype}

\usepackage{inconsolata}

\usepackage{graphicx}
\usepackage{amsmath}
\usepackage{enumerate}
\usepackage{array}
\usepackage{booktabs}
\usepackage{multirow}
\usepackage{subcaption}
\usepackage{tcolorbox}
\usepackage{listings}
\usepackage{amssymb}
\usepackage{amsthm}
\usepackage{tikz}
\usepackage{algorithm}
\usepackage{algpseudocode}
\usepackage{bm}
\usepackage{tabularx}
\usepackage{newfloat}
\usepackage{longtable}
\usepackage{geometry}

\title{
Preference Curriculum: \\
LLMs Should Always Be Pretrained on Their Preferred Data
}
\author{
    Xuemiao Zhang\textsuperscript{1,4}$^{\ast}$, \ 
    Liangyu Xu\textsuperscript{4}$^{\ast}$, \ 
    Feiyu Duan\textsuperscript{2,4}\thanks{Equal contribution.}, \\ \ 
    \textbf{Yongwei Zhou\textsuperscript{4}}$^{\ast}$, \ 
    \textbf{Sirui Wang\textsuperscript{3,4}\thanks{Corresponding author.}}, \
     \textbf{Rongxiang Weng\textsuperscript{4}}, \
    \textbf{Jingang Wang\textsuperscript{4}}, \ 
    \textbf{Xunliang Cai\textsuperscript{4}}
    \\
    \textsuperscript{1} Peking University\quad
    \textsuperscript{2} Beihang University\quad
    \textsuperscript{3} Tsinghua University\quad
    \textsuperscript{4} Meituan \\
    \texttt{zhangxuemiao@pku.edu.cn}\quad 
    \texttt{duanfeiyu@buaa.edu.cn} \quad \texttt{ywzhouphd2018@gmail.com}\\
    \texttt{\{xuliangyu02, wangsirui, wangjingang02, caixunliang\}@meituan.com} \\
  } 

\begin{document}
\maketitle
\begin{abstract}
Large language models (LLMs) generally utilize a consistent data distribution throughout the pretraining process. However, as the model's capability improves, it is intuitive that its data preferences dynamically change, indicating the need for pretraining with different data at various training stages. To achieve it, we propose the \textbf{P}erplexity \textbf{D}ifference (PD) based \textbf{P}reference \textbf{C}urriculum learning (\textbf{PDPC}) framework, which always perceives and uses the data preferred by LLMs to train and boost them. First, we introduce the PD metric to quantify the difference in how challenging a sample is for weak versus strong models. Samples with high PD are more challenging for weak models to learn and are more suitable to be arranged in the later stage of pretraining. Second, we propose the preference function to approximate and predict the data preference of the LLM at any training step, so as to complete the arrangement of the dataset offline and ensure continuous training without interruption.
Experimental results on 1.3B and 3B models demonstrate that PDPC significantly surpasses baselines. Notably, the 3B model trained on 1T tokens achieves an increased average accuracy of over \textbf{8.1\%} across MMLU and CMMLU.
\end{abstract}

\section{Introduction}
\label{sec:intro}
Large language models (LLMs) have shown impressive performance on various tasks after being pretrained on vast amounts of data \cite{touvron2023llama,dubey2024llama,liu2024deepseek}. As LLMs undergo extensive pretraining, their capabilities steadily improve, which influences their performance and data preferences \cite{yu2024mates}. Existing methods of uniformly sampling data throughout the pretraining process are suboptimal because they overlook the model's evolving data preferences \cite{wettig2024qurating,abbas2023semdedup,sachdeva2024train}.

Recently, some research has shifted the focus to data influence on model capability during pretraining \cite{evans2024datacurationjointexample,yu2024mates,koh2017understanding,ko2024mirrored}. A typical method named MATES considers the changing data influence on models but requires interrupting the training process to select more preferred data based on the model's current state, which disrupts training continuity and stability \cite{yu2024mates}. A similar issue arises in JEST \cite{evans2024datacurationjointexample}.

In this paper, we introduce perplexity \cite{1976Time} difference (PD) to quantify the difference in how challenging a sample is for LLMs' final and early checkpoints and use PD to gain a deeper understanding of model characteristics.
We further propose a novel \textbf{PD}-based \textbf{P}reference \textbf{C}urriculum learning framework, PDPC, as shown in Figure \ref{PDPC_framework}. It perceives LLMs' data preference at any training step and uses the preferred data to continuously pretrain LLMs without interruption, thereby boosting their performance. To this end, PDPC needs to solve three major challenges:
\begin{figure}[t]
    \centering    
    \includegraphics[width=\linewidth]{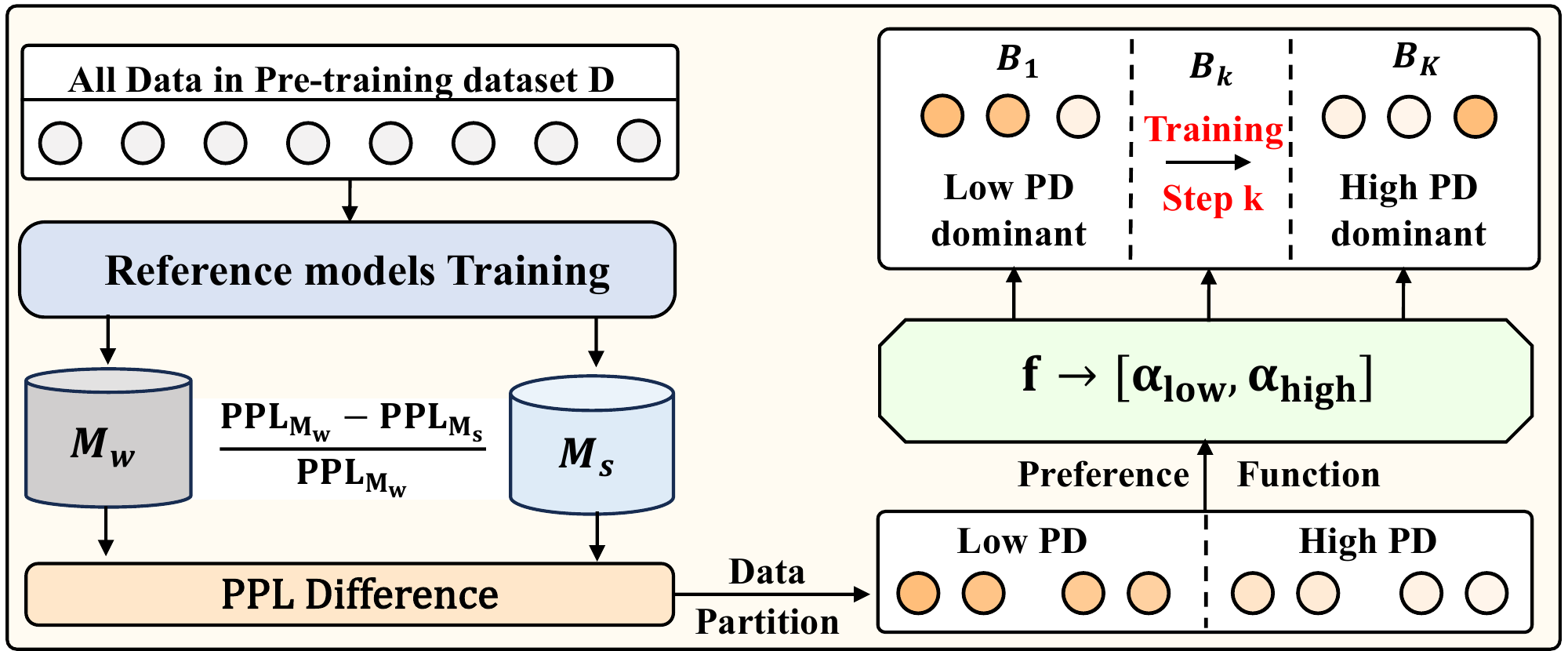}
    \caption{PD-based Preference Curriculum Framework.}
    \label{PDPC_framework}
\end{figure}
\paragraph{How to dynamically perceive preferences without interrupting pretraining.} 
Ideally, training would be paused at any training step to calculate PD using the model's current state, allowing for data sampling from the preference distribution like MATES \cite{yu2024mates}. However, to ensure continuity and stability in pretraining, we propose an offline processing method to approximate the ideal dynamic preference adjustment. First, we calculate PD for all samples offline using both the fully trained checkpoint and an early checkpoint of the experimental model. Then, we develop a preference function to predict the model's preference for data with specific PD characteristics, enabling the pre-organization of data according to the model's preferences at various training stages. Because the data is fully arranged offline, the training of the experimental model proceeds without interruption. 

\paragraph{Calculating PD for the entire dataset is prohibitively expensive.}  
Typically, the sizes of experimental models and the pretraining dataset are quite large. Training an experimental model using the entire dataset in a random sampling setting is very costly. Moreover, calculating PD values offline necessitates inferring the entire dataset twice—using both early and final checkpoints—which is extremely costly. Essentially, the differences between various model states can be captured by the disparity in trained FLOPs: the early checkpoint corresponds to fewer trained FLOPs, whereas the final checkpoint corresponds to more. Naturally, we can approximate the FLOPs difference by using two reference models (RMs) with fewer parameters compared to the experimental model, pretraining them on the same data scale. The smaller and larger RM play the roles of the early checkpoint and the final checkpoint, respectively. This approach significantly reduces both pretraining and inference costs.

\paragraph{How to use PD to orderly arrange pretraining data.} 
Data with high PD are challenging for weak models but well-suited for strong models due to the increased capacity. Conversely, data with low PD are less sensitive to model capability differences and can be understood by both strong and weak models. From the perspective of FLOPs, the early stages of training an experimental model can be seen as involving a weak, smaller model, while later stages involve a strong, larger model. Thus, placing high-PD data in the later training stages can enhance the model's ability to fit these challenging samples, whereas low-PD data can be trained earlier since they are less sensitive to model capability. This establishes a natural curriculum learning principle for the pretraining: begin with low-PD data and progress to high-PD data.  Following the principle, we propose an S-shape function that effectively models preferences throughout training. Further, to maintain diversity in each batch, we use the concentration of low-PD data rather than PD itself as the output of the preference function.

By overcoming the challenges, PDPC can always perceive and use the data preferred by LLMs to pretrain and boost them. Notably, PDPC only arranges the given data without performing data selection. In summary, our work has the following contributions:
(1) We propose PD to measure the difference in the fit of strong and weak models to samples, pointing out that high-PD data is challenging for weak models and is suitable to be arranged in the later pretraining stage.
(2) We propose a novel PDPC framework that always perceives and uses the data preferred by LLMs to train and boost them, and ensures uninterrupted continuous training, which serves as the last data preprocessing step.
(3) Experimental results show significant performance improvements over baselines. Notably, the 3B model trained on 1T tokens with PDPC demonstrates an average accuracy increase of \textbf{4.1\%} across all benchmarks and \textbf{8.1\%} across MMLU and CMMLU.

\section{PD-based Preference Curriculum} 
In this section, we propose PDPC, which always perceives and uses the data preferred by LLMs to pretrain and boost them. Notably, PDPC only organizes the given data without performing selection, which can serve as the final data preprocessing step before pretraining.

\subsection{Problem Formulation} \label{subsec:probdef}
Given a pretraining dataset \(\mathcal{D}\) following a uniform distribution, we aim to arrange it into a form that better aligns with the oracle distribution \(\mathcal{O}_{\{B_k\}}\), which represents the ideal data distribution preferred by the model. To achieve this, we aim to find the optimal preference function \(f^*\) that adjusts the joint distribution of all batches \(\{B_k\}_{k=1}^K\) to closely approximate \(\mathcal{O}_{\{B_k\}}\).
\begin{equation}
\begin{aligned}
f^* = & \arg\min_{f} \,  \text{Divergence}\left(\mathcal{P}_{\{B_k\}}, \mathcal{O}_{\{B_k\}}\right) \\
&\text{s.t.} \quad  \bigcup_{k}^K B_k = \mathcal{D}, ~ B_k = f\left(\frac{k}{K}\right)
\end{aligned}
\end{equation}
where \(\mathcal{P}_{\{B_k\}}\) represents the joint distribution of the batches, and \(K\) is the total number of batches.

\subsection{PD-based Data Partitioning}
\paragraph{Perplexity Difference (PD).} We begin by introducing the concept of PD. Consider two models, the weak model $M_w$ and the strong model $M_s$, both trained on an identical dataset $\mathcal{D}$. Given a sample \( x \), the PD is defined as:
\begin{equation}
\begin{aligned}
    &PD(x) = \frac{PPL_{M_w}(x) - PPL_{M_s}(x)}{PPL_{M_w}(x)} \\
    &PPL_{M_{*}}(x) = exp{(-\frac{1}{L_x}\sum_{t=1}^{L_x}\log P(x_t|x_{<t}))}
\end{aligned}
\end{equation}
where \( PPL_{M_w}(x) \) and \( PPL_{M_s}(x) \) are the perplexity values of the sample \( x \) calculated using \( M_w \) and \( M_s \), respectively. \( L_x \) denotes the token length of the sample \( x \), and \( x_t \) represents the \( t \)-th token. "*" indicates weak or strong. 

PD indicates the extent to which the strong model outperforms the weak model on a given sample. Low PD indicates that the strong model \( M_s \) and the weak model \( M_w \) have a similar level of fit to the sample \( x \). Conversely, a high PD value suggests that \( M_s \) outperforms \( M_w \) in fitting the sample \( x \), implying that the sample is more challenging for \( M_w \) to learn. There are very few samples whose perplexity in \(M_w\) is smaller than in \(M_s\), accounting for less than 0.01\%, and we ignore them.

\paragraph{Data Partitioning.} An intuitive method for data arrangement is to sort the pretraining data by PD from low to high. However, this method is suboptimal. As shown in Figure \ref{data_distribution} and \ref{combined_figures2} of Section \ref{sec:case}, data with extremely high and extremely low PD values differ significantly. Sorting by PD creates batches with overly homogeneous samples, lacking the diversity essential for pretraining LLMs \cite{sachdeva2024train}. 

To solve this issue, we propose to partition data into distinct parts based on mutually exclusive PD ranges. 
During each pretraining step, data from each part is mixed in varying proportions. Formally, given a pretraining dataset \(\mathcal{D}\), we calculate the PD for each data point \(x\), denoted as \(\text{PD}_x\). We then sort \(\mathcal{D}\) in ascending order based on \(\text{PD}_x\) and partition it evenly into \(n\) parts, \(\mathcal{D}_1, \mathcal{D}_2, \ldots, \mathcal{D}_n\): 
\begin{equation}
   |\mathcal{D}_i| = \frac{|\mathcal{D}|}{n} \quad \forall i \in \{1, 2, \ldots, n\}
\end{equation}
For any \(i < j\), the condition holds that:
\begin{equation}
    \forall x \in \mathcal{D}_i, \forall y \in \mathcal{D}_j, \quad \text{PD}_x \leq \text{PD}_y
\end{equation}

\subsection{CL with PD-based Preference Function}
\label{sec:2.4}
We introduce a preference function $f(p)$ to capture the model's preferences for different PD partitions at any training step. 
The function maps the pretraining progress \( p \) to a proportion vector \( \boldsymbol{\alpha} = [\alpha_1, \alpha_2, \ldots, \alpha_n] \), where \( \alpha_i \) denotes the fraction of data from the \(i\)-th domain in the current batch. The training completion rate \( p \) is defined as: $p = \frac{k}{K}$, where \( k \) is the current training step and \( K \) is the total number of training steps. The function \( f(p) \) is defined as 
\begin{equation}
\begin{aligned}
    f&(p) \rightarrow [\alpha_1(p), \alpha_2(p), \ldots, \alpha_n(p)]
    \\ &\text{s.t.} \sum_{i=1}^{n} \alpha_i(p) = 1 
\end{aligned}
\end{equation}
which determines how the proportion of each part changes as training progresses.

To ensure full utilization of data from all parts by the end of the training, we establish the constraint \(\int_{0}^{1} \alpha_i(p) \, dp = \frac{1}{n}\) for all \(i \in \{1, 2, \ldots, n\}\). In each training step, the current proportion vector \(\boldsymbol{\alpha}\) guides the proportional selection of samples from each part \(\mathcal{D}_i\). These selected samples are then combined to form a new mixed batch, which is used for the training step.

Following the CL principle discussed in Section \ref{sec:intro}, \textbf{PDPC starts the training process using low-PD data and gradually moves to high-PD data.} To implement this, we sort the pretraining data and partition them into \( n \) parts, allowing us to explore the model's preference for mixing ratios of data with different PD values at various pretraining steps. We discuss different scenarios:

\noindent(1) When \(n=1\), the training process involves randomly selecting individual samples from the dataset for each training step.
    
\noindent(2) When \( {n=|\mathcal{D}|} \), each part corresponds to a single sample, which is equivalent to performing a full sample-level sorting of all the data.
    
\noindent(3) When \({1 < n < |\mathcal{D}|}\), each training step will include data from different parts. Finding the optimal function \( f^\star \) is a complex problem that involves substantial costs.


We focus on the case where \(n=2\) because it effectively captures the essential differences between low-PD and high-PD data while remaining computationally manageable. Essentially, \(n=2\) partitions the data into high-PD and low-PD parts. 

To guide the offline organization of pretraining data, We introduce a PD-based preference function \(f(\frac{k}{K})\), which predicts the proportion of low PD data that the model prefers at different training steps $k$. However, directly optimizing the preference function \(f\) is challenging. To solve this issue, we propose a function search method\footnote{We also discuss an annealing-based iterative optimization approach in Appendix \ref{sec:apc} to approximate $f^*$.} to approximate $f^*$. To ensure equal volumes of high and low PD data and align with the model's data preferences, it is crucial to choose the right function. This function ensures that the proportion \( b \) of low PD data matches the pretraining completion \( p = \frac{k}{K} \), expressed as \( b = f(p) \), and must meet specific criteria:

\noindent \textbf{Firstly}, based on the CL principle, the function \( f(p) \) should exhibit a decreasing trend, gradually increasing the proportion of data with high PD to raise the curriculum difficulty.

\noindent \textbf{Secondly}, to ensure that the total amount of data with high PD and low PD stays equal, the function can be symmetric about the point \((0.5, 0.5)\), satisfying \(f(0.5 + \Delta) = 1 - f(0.5 - \Delta)\), where \(\Delta \in (0, 0.5)\).
\begin{figure}[ht]
    \centering
    \begin{subfigure}[b]{0.49\linewidth}
        \includegraphics[width=\linewidth]{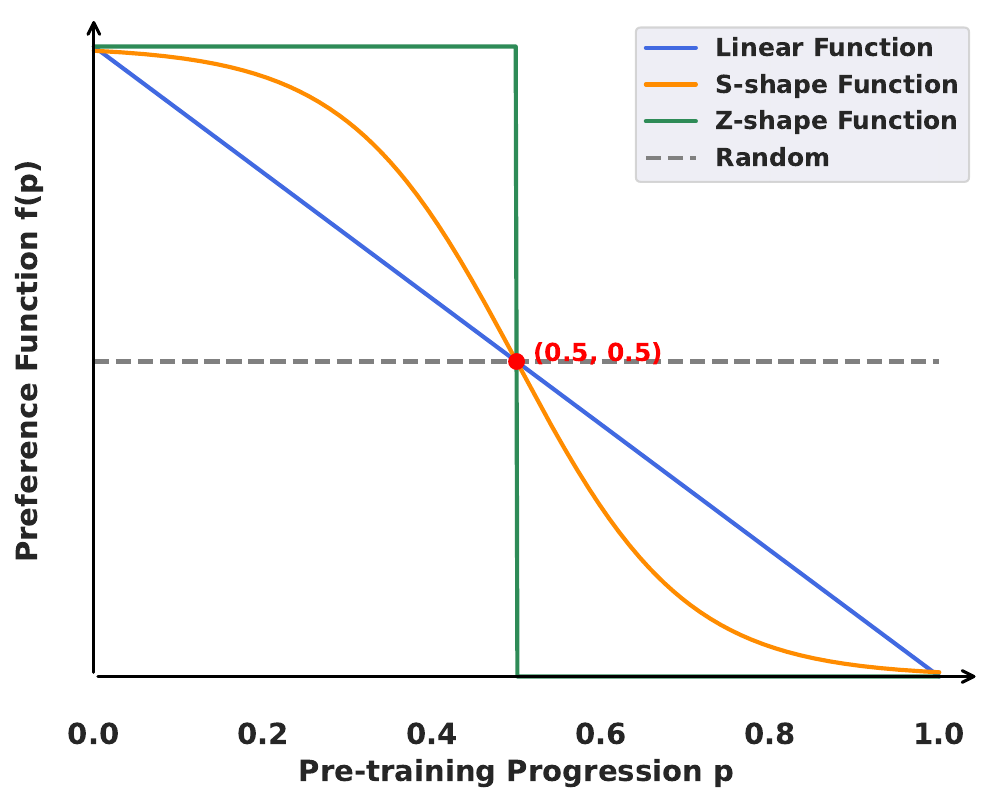}
        \label{fig:function}
    \end{subfigure}
    \hfill
    \begin{subfigure}[b]{0.49\linewidth}
    \includegraphics[width=\linewidth]{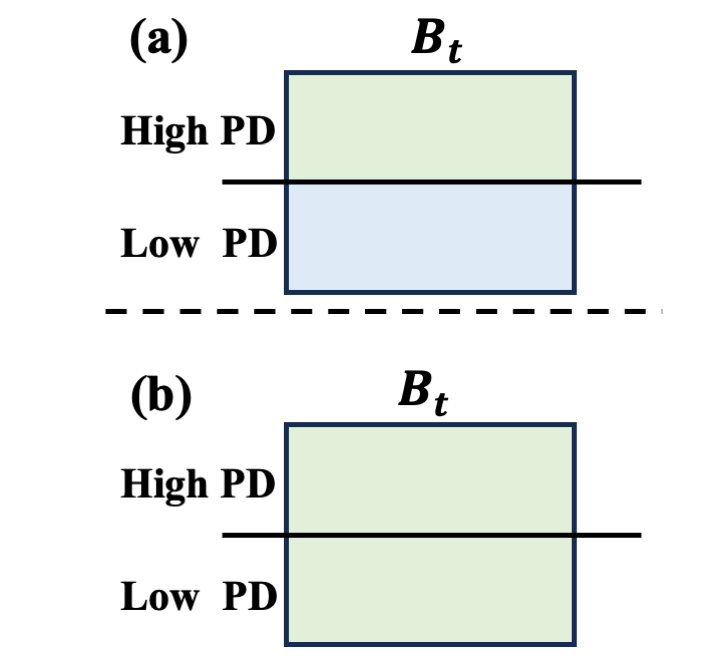}
        \label{fig:pooling}
    \end{subfigure}
    \caption{Left: Preference Functions and Right: a comparison of different data sampling methods, the regions highlighted in green represent the preferred data.}
    \label{fig_no}
\end{figure}

We initially search for three representative functions as candidates: the linear function $f_{L}(p)$, the Z-shape function $f_{Z}(p)$, and the S-shape function $f_{S}(p)$, as shown in the left part of Figure \ref{fig_no}. $f_{L}(p)$ represents a steady decline, while $f_{Z}(p)$ indicates a sudden, distinct change. To introduce a balance between the two, we also incorporate an S-shape function form.
\begin{equation}
   f_{L}(p) = k \cdot (p - 0.5) + 0.5
   \label{eq:linear}
\end{equation}  
\begin{equation}
   f_{Z}(p) = 
\begin{cases} 
1 - \lambda, & \text{if } p < 0.5 \\
\lambda, & \text{if } p \geq 0.5 
\end{cases}
\label{eq:step}
\end{equation}
\begin{equation}
       f_{S}(p) = \frac{1}{1 + exp{(a(p - 0.5))}}
       \label{eq:sigmoid}
\end{equation}
where $k \in \left[-1,0\right), \lambda \in \left[0, 0.5\right)$, \( a \) modulates the steepness of the curve.

Empirical results show that the S-shape function, with appropriate parameter tuning, effectively captures the model's preference for data proportions at different pretraining stages, eliminating the need to explore more complex function forms. 



Essentially, assuming that at step \(i\) the model prefers high PD data, traditional methods (Figure \ref{fig_no}(a)) dilute the benefits of high PD data with less favorable low PD data, similar to average pooling. In contrast, PDPC (Figure \ref{fig_no}(b)) clusters preferred data within each batch, resembling the effect of Max-Pooling on this data.


\begin{algorithm}[t]
\caption{\small PD-based Preference Curriculum Learning}
\small
\begin{algorithmic}[1]
\State \textbf{Input:} dataset $\mathcal{D}$, total iterations $K$, batch size $N$
\State \textbf{Output:} trained model $\theta_K$
\State Initialize model parameters $\theta_0$
\State Train RMs on i.i.d. subset of $\mathcal{D}$ 
\State Calculate PD: for all samples in $D$ using RMs
\State Partition \( \mathcal{D} \) into \( 2 \) sub-domains \( A_{PD}^{low} \) and $A_{PD}^{high}$.  
\State Explore and determine the form of preference function:
\State \quad \quad $f(p) = \frac{1}{1 + exp{(a(p - 0.5))}}$
\For{$k = 0$ \textbf{to} $K-1$}
    \State Calculate pretraining progress  $p = \frac{k}{K}$
    \State Get the proportions vector: 
    \State  \quad  $[\alpha_{1}, \alpha_{2}] \gets [f(p), 1-f(p)]$
    \State Sample from the two domains to form $B_{k}$:
    \State $B_{k} = \{x \mid x \sim A_{PD}^{low}\}_{\alpha_{1} \cdot N} \cup \{x \mid x \sim A_{PD}^{high}\}_{\alpha_{2} \cdot N}$
    \State Train the model on \( B_k \) and update \(\theta_k\)
\EndFor

\end{algorithmic}
\end{algorithm}

\section{Experiments}

\subsection{Experimental Setup}

\paragraph{General Setting} We train two experimental models: a 1.3B model using 100B randomly selected tokens from the SlimPajama dataset \cite{cerebras2023slimpajama}, and a 3B model on a bilingual dataset containing 1T tokens, which comprises 500B tokens each of Chinese and English data, sourced from domains such as books\cite{gao2020pile}, blogs\cite{baumgartner2020pushshift}, patents\cite{sharma2019bigpatent}, Common Crawl\cite{penedo2024fineweb}, and Wikipedia, similar to the Matrix dataset\cite{zhang2024mapneo}. We train 100M and 700M reference models (RMs) on an i.i.d. subset with 50B tokens from SlimPajama for the 1.3B setting, and 100M and 1.3B RMs on an i.i.d. subset with 500B tokens for the 3B setting, respectively. All models were trained using the Llama architecture\cite{touvron2023llama} within the Megatron framework \cite{shoeybi2019megatron}, utilizing the Adam optimizer. We set the batch size to 640 and the context window length to 8192. The initial learning rate is set to 2e-4, with a warm-up phase of over 375M tokens. We adopt a cosine learning rate schedule and set weight decay to 0.1. A full pretraining run of the 3B model on 1 trillion tokens, utilizing 512 Ascend 910B NPUs, requires approximately 180 hours.

\paragraph{Evaluation} We employ the lm-evaluation-harness \cite{gao2021framework} to measure the models' performance on the following benchmarks: ARC-E \cite{clark2018think}, ARC-C \cite{clark2018think}, SciQ \cite{welbl2017crowdsourcing}, HellaSwag \cite{zellers2019hellaswag} and PIQA \cite{bisk2020piqa}, which include tasks like knowledge question answering and commonsense reasoning. For the 3B model, we add benchmarks like MMLU \cite{hendrycks2020measuring}, CMMLU\cite{li2024cmmlumeasuringmassivemultitask}, and CEVAL\cite{huang2023cevalmultilevelmultidisciplinechinese}, which cover multi-domain knowledge and complex reasoning tasks, presenting challenges absent in 1.3B models.  We employ in-context learning for evaluation following QuRating \cite{wettig2024qurating}. Standard accuracy is used as the final metric for all tasks.

\paragraph{Baselines} We compare PDPC with \textit{Random} and several basic curriculum learning approaches:
\textbf{(1) \textit{Random}}: Each batch is randomly selected from the entire dataset, corresponding to the case of \(n=1\) in our framework.
\textbf{(2) \textit{PPL}}: PPL directly measures how well a model fits the data. We use two 700M models, each trained separately on the SlimPajama and Matrix subsets, to annotate their corresponding data.
\textbf{(3) \textit{QuRating}} \cite{wettig2024qurating}:  We select the Education Value in QuRating as the difficulty indicator for curriculum learning.
\textbf{(4) \textit{Sequential}}: We fully sort the data based on PD, PPL, and Qurating, arranging them in either ascending or descending order.

\subsection{Main Results}
\begin{table*}[t]
    \centering
    \setlength{\fboxsep}{1pt} 
    \scalebox{0.9}{
    \small
    \begin{tabular}{c|
    >{\centering\arraybackslash}m{1.5cm} 
    >{\centering\arraybackslash}m{1.5cm}|
    >{\centering\arraybackslash}m{1.5cm} 
    >{\centering\arraybackslash}m{1.5cm}
    >{\centering\arraybackslash}m{1cm}
    >{\centering\arraybackslash}m{1.2cm}
    >{\centering\arraybackslash}m{1cm}
    >{\centering\arraybackslash}m{1cm}
    }
    \toprule
        \textbf{Method} & \textbf{Metric} & \textbf{Order} & \textbf{ARC-E} & \textbf{ARC-C} & \textbf{SciQ} & \textbf{HellaSw.} & \textbf{PIQA} & \textbf{AVG.} \\ 
    \midrule
        \multicolumn{1}{c}{-} & - & Random & 56.5 & 23.6 & 85.8 & 34.2 & 67.3 & 53.5 \\ 
    \midrule
        \multirow{6}{*}{Sequential} & PD & High2Low & 54.7\textsuperscript{\hspace{0.3em}\raisebox{0.5ex}{\colorbox{red!10}{\tiny$\downarrow$1.8}}} & 21.8\textsuperscript{\hspace{0.3em}\raisebox{0.5ex}{\colorbox{red!10}{\tiny$\downarrow$1.8}}} & 87.1\textsuperscript{\hspace{0.3em}\raisebox{0.5ex}{\colorbox{green!10}{\tiny$\uparrow$1.3}}} & 33.7\textsuperscript{\hspace{0.3em}\raisebox{0.5ex}{\colorbox{red!10}{\tiny$\downarrow$0.5}}} & 67.8\textsuperscript{\hspace{0.3em}\raisebox{0.5ex}{\colorbox{green!10}{\tiny$\uparrow$0.5}}} & 53.0\textsuperscript{\hspace{0.3em}\raisebox{0.5ex}{\colorbox{red!10}{\tiny$\downarrow$0.5}}} \\
        & PD & Low2High & 56.1\textsuperscript{\hspace{0.3em}\raisebox{0.5ex}{\colorbox{red!10}{\tiny$\downarrow$0.4}}} & 21.3\textsuperscript{\hspace{0.3em}\raisebox{0.5ex}{\colorbox{red!10}{\tiny$\downarrow$2.3}}} & 86.2\textsuperscript{\hspace{0.3em}\raisebox{0.5ex}{\colorbox{green!10}{\tiny$\uparrow$0.4}}} & 34.4\textsuperscript{\hspace{0.3em}\raisebox{0.5ex}{\colorbox{green!10}{\tiny$\uparrow$0.2}}} & 67.6\textsuperscript{\hspace{0.3em}\raisebox{0.5ex}{\colorbox{red!10}{\tiny$\downarrow$0.2}}} & 53.1\textsuperscript{\hspace{0.3em}\raisebox{0.5ex}{\colorbox{red!10}{\tiny$\downarrow$0.4}}} \\
        & PPL & High2Low & 45.5\textsuperscript{\hspace{0.3em}\raisebox{0.5ex}{\colorbox{red!10}{\tiny$\downarrow$11.0}}} & 20.6\textsuperscript{\hspace{0.3em}\raisebox{0.5ex}{\colorbox{red!10}{\tiny$\downarrow$3.0}}} & 71.2\textsuperscript{\hspace{0.3em}\raisebox{0.5ex}{\colorbox{red!10}{\tiny$\downarrow$14.6}}} & 30.3\textsuperscript{\hspace{0.3em}\raisebox{0.5ex}{\colorbox{red!10}{\tiny$\downarrow$3.9}}} & 63.7\textsuperscript{\hspace{0.3em}\raisebox{0.5ex}{\colorbox{red!10}{\tiny$\downarrow$3.6}}} & 46.3\textsuperscript{\hspace{0.3em}\raisebox{0.5ex}{\colorbox{red!10}{\tiny$\downarrow$7.2}}} \\
        & PPL & Low2High & 47.8\textsuperscript{\hspace{0.3em}\raisebox{0.5ex}{\colorbox{red!10}{\tiny$\downarrow$8.7}}} & 17.9\textsuperscript{\hspace{0.3em}\raisebox{0.5ex}{\colorbox{red!10}{\tiny$\downarrow$5.7}}} & 72.7\textsuperscript{\hspace{0.3em}\raisebox{0.5ex}{\colorbox{red!10}{\tiny$\downarrow$13.1}}} & 29.1\textsuperscript{\hspace{0.3em}\raisebox{0.5ex}{\colorbox{red!10}{\tiny$\downarrow$5.1}}} & 62.4\textsuperscript{\hspace{0.3em}\raisebox{0.5ex}{\colorbox{red!10}{\tiny$\downarrow$4.9}}} & 46.0\textsuperscript{\hspace{0.3em}\raisebox{0.5ex}{\colorbox{red!10}{\tiny$\downarrow$7.5}}} \\
        & Qu.Edu & High2Low & 57.2\textsuperscript{\hspace{0.3em}\raisebox{0.5ex}{\colorbox{green!10}{\tiny$\uparrow$0.7}}} & 26.4\textsuperscript{\hspace{0.3em}\raisebox{0.5ex}{\colorbox{green!10}{\tiny$\uparrow$2.8}}} & 85.4\textsuperscript{\hspace{0.3em}\raisebox{0.5ex}{\colorbox{red!10}{\tiny$\downarrow$0.4}}} & 33.0\textsuperscript{\hspace{0.3em}\raisebox{0.5ex}{\colorbox{red!10}{\tiny$\downarrow$1.2}}} & 66.2\textsuperscript{\hspace{0.3em}\raisebox{0.5ex}{\colorbox{red!10}{\tiny$\downarrow$1.1}}} & 53.6\textsuperscript{\hspace{0.3em}\raisebox{0.5ex}{\colorbox{green!10}{\tiny$\uparrow$0.1}}} \\
        & Qu.Edu & Low2High & 56.8\textsuperscript{\hspace{0.3em}\raisebox{0.5ex}{\colorbox{green!10}{\tiny$\uparrow$0.3}}} & 26.0\textsuperscript{\hspace{0.3em}\raisebox{0.5ex}{\colorbox{green!10}{\tiny$\uparrow$2.4}}} & 84.1\textsuperscript{\hspace{0.3em}\raisebox{0.5ex}{\colorbox{red!10}{\tiny$\downarrow$1.7}}} & 33.5\textsuperscript{\hspace{0.3em}\raisebox{0.5ex}{\colorbox{red!10}{\tiny$\downarrow$0.7}}} & 67.9\textsuperscript{\hspace{0.3em}\raisebox{0.5ex}{\colorbox{green!10}{\tiny$\uparrow$0.6}}} & 53.7\textsuperscript{\hspace{0.3em}\raisebox{0.5ex}{\colorbox{green!10}{\tiny$\uparrow$0.2}}} \\
    \midrule
        \multirow{5}{*}{Preference CL} & PPL & S.R. & 56.1\textsuperscript{\hspace{0.3em}\raisebox{0.5ex}{\colorbox{red!10}{\tiny$\downarrow$0.4}}} & 24.1\textsuperscript{\hspace{0.3em}\raisebox{0.5ex}{\colorbox{green!10}{\tiny$\uparrow$0.5}}} & 87.8\textsuperscript{\hspace{0.3em}\raisebox{0.5ex}{\colorbox{green!10}{\tiny$\uparrow$2.0}}} & 33.9\textsuperscript{\hspace{0.3em}\raisebox{0.5ex}{\colorbox{red!10}{\tiny$\downarrow$0.3}}} & 67.4\textsuperscript{\hspace{0.3em}\raisebox{0.5ex}{\colorbox{green!10}{\tiny$\uparrow$0.1}}} & 53.9\textsuperscript{\hspace{0.3em}\raisebox{0.5ex}{\colorbox{green!10}{\tiny$\uparrow$0.4}}} \\
        & PPL & S. & 56.1\textsuperscript{\hspace{0.3em}\raisebox{0.5ex}{\colorbox{red!10}{\tiny$\downarrow$0.4}}} & 22.6\textsuperscript{\hspace{0.3em}\raisebox{0.5ex}{\colorbox{red!10}{\tiny$\downarrow$1.0}}} & 85.5\textsuperscript{\hspace{0.3em}\raisebox{0.5ex}{\colorbox{red!10}{\tiny$\downarrow$0.3}}} & \textbf{34.2}\textsuperscript{\hspace{0.3em}\raisebox{0.5ex}{{\tiny ~~~0.0}}} & 67.5\textsuperscript{\hspace{0.3em}\raisebox{0.5ex}{\colorbox{green!10}{\tiny$\uparrow$0.2}}} & 53.2\textsuperscript{\hspace{0.3em}\raisebox{0.5ex}{\colorbox{red!10}{\tiny$\downarrow$0.3}}} \\
        & Qu.Edu & S.R & 56.7\textsuperscript{\hspace{0.3em}\raisebox{0.5ex}{\colorbox{green!10}{\tiny$\uparrow$0.2}}} & 24.9\textsuperscript{\hspace{0.3em}\raisebox{0.5ex}{\colorbox{green!10}{\tiny$\uparrow$1.3}}} & 86.2\textsuperscript{\hspace{0.3em}\raisebox{0.5ex}{\colorbox{green!10}{\tiny$\uparrow$0.4}}} & 33.6\textsuperscript{\hspace{0.3em}\raisebox{0.5ex}{\colorbox{red!10}{\tiny$\downarrow$0.6}}} & 66.9\textsuperscript{\hspace{0.3em}\raisebox{0.5ex}{\colorbox{red!10}{\tiny$\downarrow$0.4}}} & 53.7\textsuperscript{\hspace{0.3em}\raisebox{0.5ex}{\colorbox{green!10}{\tiny$\uparrow$0.2}}} \\
        & Qu.Edu & S. & 55.5\textsuperscript{\hspace{0.3em}\raisebox{0.5ex}{\colorbox{red!10}{\tiny$\downarrow$1.0}}} & 24.8\textsuperscript{\hspace{0.3em}\raisebox{0.5ex}{\colorbox{green!10}{\tiny$\uparrow$1.2}}} & 87.8\textsuperscript{\hspace{0.3em}\raisebox{0.5ex}{\colorbox{green!10}{\tiny$\uparrow$2.0}}} & 34.0\textsuperscript{\hspace{0.3em}\raisebox{0.5ex}{\colorbox{green!10}{\tiny$\uparrow$0.2}}} & 67.4\textsuperscript{\hspace{0.3em}\raisebox{0.5ex}{\colorbox{green!10}{\tiny$\uparrow$0.1}}} & 53.9\textsuperscript{\hspace{0.3em}\raisebox{0.5ex}{\colorbox{green!10}{\tiny$\uparrow$0.4}}} \\
        & PD & S.R. & 56.7\textsuperscript{\hspace{0.3em}\raisebox{0.5ex}{\colorbox{green!10}{\tiny$\uparrow$0.2}}} & 24.9\textsuperscript{\hspace{0.3em}\raisebox{0.5ex}{\colorbox{green!10}{\tiny$\uparrow$1.3}}} & 86.2\textsuperscript{\hspace{0.3em}\raisebox{0.5ex}{\colorbox{green!10}{\tiny$\uparrow$0.4}}} & 33.6\textsuperscript{\hspace{0.3em}\raisebox{0.5ex}{\colorbox{red!10}{\tiny$\downarrow$0.6}}} & 67.4\textsuperscript{\hspace{0.3em}\raisebox{0.5ex}{\colorbox{green!10}{\tiny$\uparrow$0.1}}} & 53.8\textsuperscript{\hspace{0.3em}\raisebox{0.5ex}{\colorbox{green!10}{\tiny$\uparrow$0.3}}} \\
    \midrule
        PDPC & PD & S. & \textbf{57.3}\textsuperscript{\hspace{0.3em}\raisebox{0.5ex}{\colorbox{green!10}{\tiny\textbf{$\uparrow$0.8}}}} & \textbf{26.6}\textsuperscript{\hspace{0.3em}\raisebox{0.5ex}{\colorbox{green!10}{\tiny\textbf{$\uparrow$2.9}}}} & \textbf{87.9}\textsuperscript{\hspace{0.3em}\raisebox{0.5ex}{\colorbox{green!10}{\tiny\textbf{$\uparrow$2.1}}}} & 33.7\textsuperscript{\hspace{0.3em}\raisebox{0.5ex}{\colorbox{red!10}{\tiny$\downarrow$0.5}}} & \textbf{68.0}\textsuperscript{\hspace{0.3em}\raisebox{0.5ex}{\colorbox{green!10}{\tiny\textbf{$\uparrow$0.7}}}} & \textbf{54.7}\textsuperscript{\hspace{0.3em}\raisebox{0.5ex}{\colorbox{green!10}{\tiny\textbf{$\uparrow$1.2}}}} \\
        
    \bottomrule
    \end{tabular}
    }
    \caption{Downstream tasks results on \textbf{1.3B} models with \textbf{100B} tokens. We report accuracy for each task, and the best performances are marked in bold. Abbreviations: HellaSw. = HellaSwag, AVG. = Average, S.=S-shape Function, S.R.=S-shape Reverse Function.}
    \label{main results1}
\end{table*}

\begin{figure}[t]
    \centering
    \begin{minipage}[b]{0.49\linewidth}
        \includegraphics[width=\linewidth]{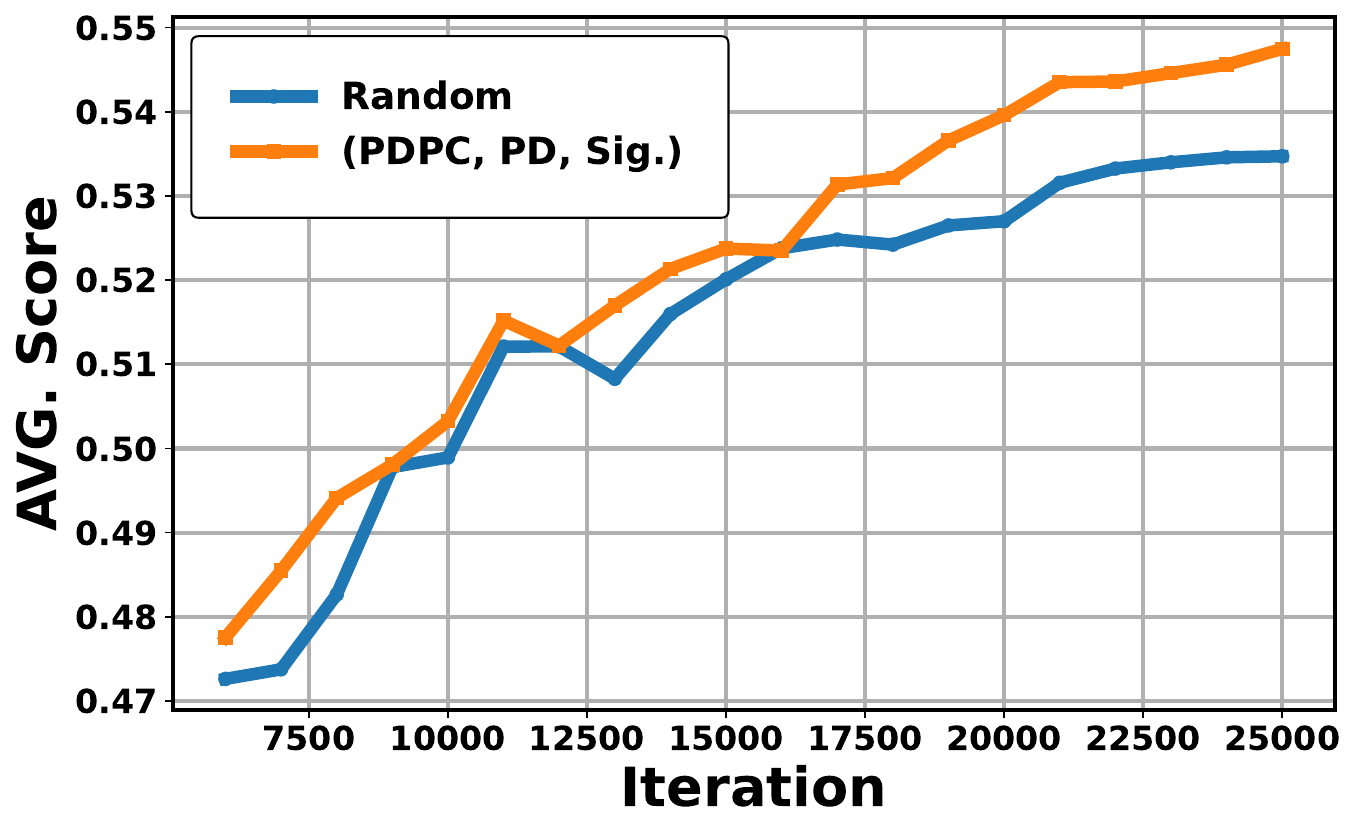}
        \caption*{(a) 1B AVG. w/ steps.}
    \end{minipage}
    \hfill
    \begin{minipage}[b]{0.49\linewidth}
        \includegraphics[width=\linewidth]{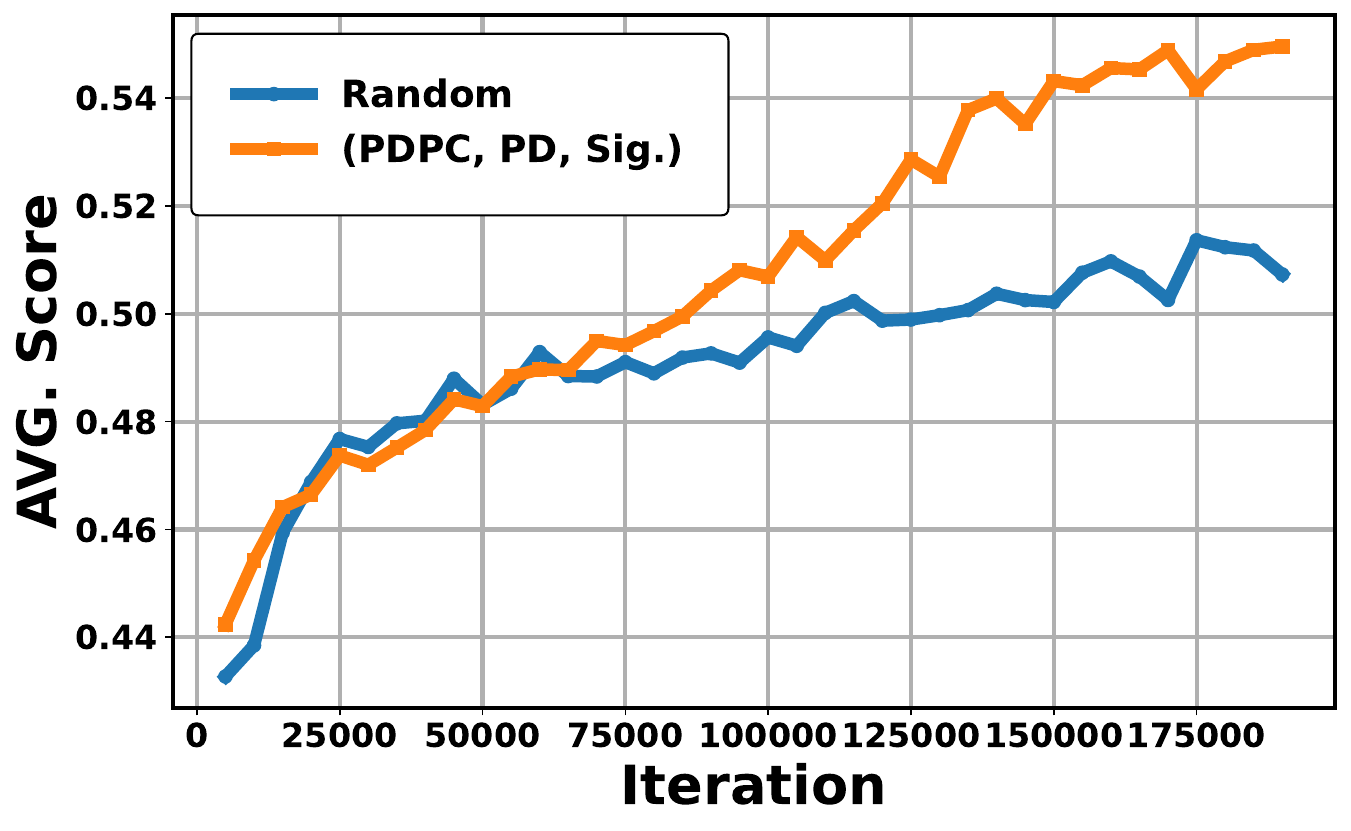}
        \caption*{(b) 3B AVG. w/ steps.}
    \end{minipage}
    \begin{minipage}[b]{0.49\linewidth}
        \includegraphics[width=\linewidth]{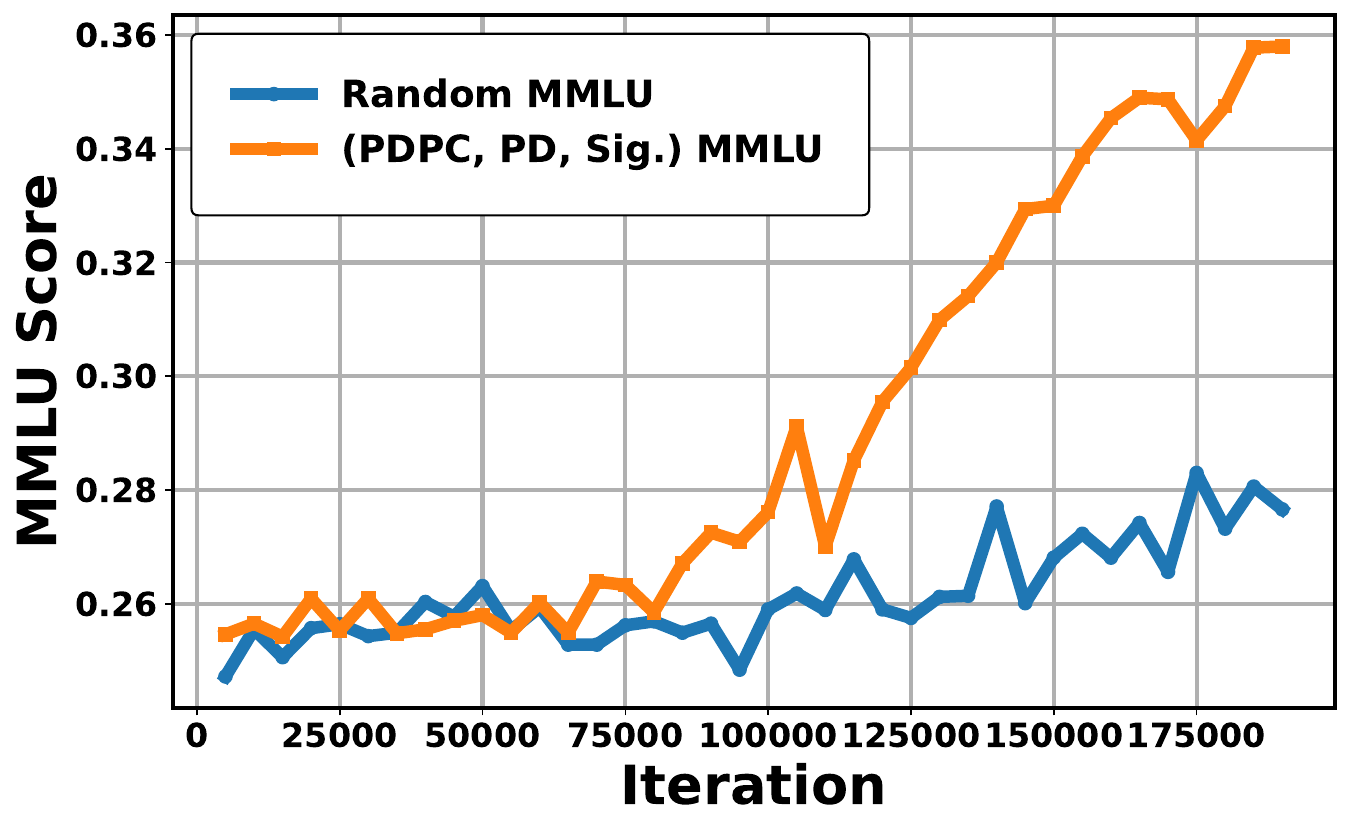}
        \caption*{(c) 3B MMLU w/ steps.}
    \end{minipage}
    \hfill
    \begin{minipage}[b]{0.49\linewidth}
        \includegraphics[width=\linewidth]{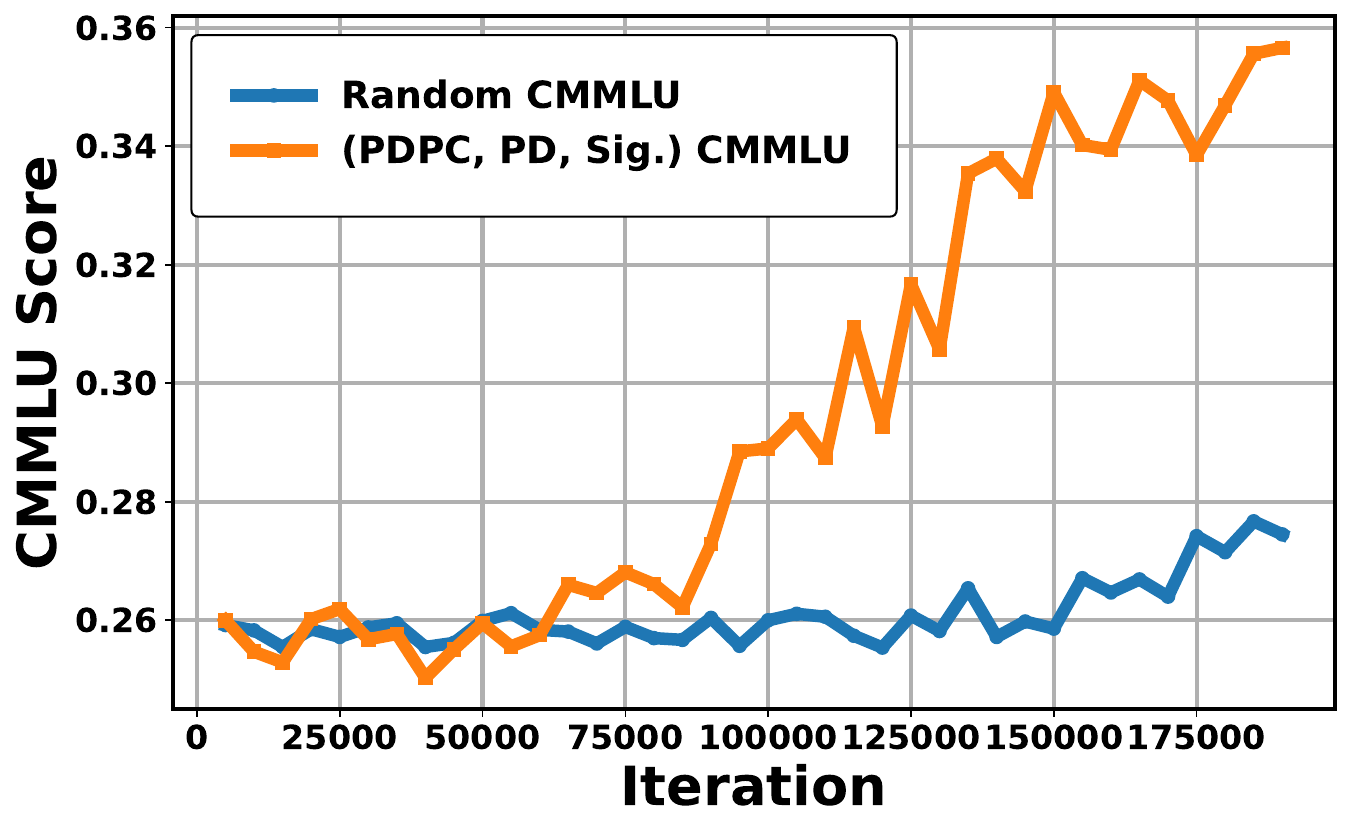}
        \caption*{(d) 3B CMMLU w/ steps.}
    \end{minipage}
    \caption{Few-shot downstream performance with respect to training steps for Random and (PDPC, PD, S.).}
    \label{fig:combinedplot}
\end{figure}

\begin{table*}[t]
    \centering
    \setlength{\fboxsep}{1pt} 
    \scalebox{0.9}{
    \small
    \begin{tabular}{c
    >{\centering\arraybackslash}m{1cm} 
    >{\centering\arraybackslash}m{1.1cm}|
    >{\centering\arraybackslash}m{1.2cm} 
    >{\centering\arraybackslash}m{1.2cm}
    >{\centering\arraybackslash}m{0.8cm}
    >{\centering\arraybackslash}m{0.8cm}
    >{\centering\arraybackslash}m{1cm}
    >{\centering\arraybackslash}m{1cm}
    >{\centering\arraybackslash}m{1cm}
    >{\centering\arraybackslash}m{1cm}
    }
    \toprule
        \textbf{Method} & \textbf{Metric} & \textbf{Order} & \textbf{ARC-E} & \textbf{ARC-C} & \textbf{SciQ}  &  \textbf{PIQA} &\textbf{MMLU} & \textbf{CMMLU} & \textbf{CEVAL} &  \textbf{AVG.} \\ 
    \midrule
        \multicolumn{1}{c}{-} & - & Random & 68.6 & 33.7 & 94.6 & 76.0 & 27.7 & 27.5 & 27.2 & 50.8 \\ 
    \midrule
        PDPC & PD & S. & \textbf{69.7}\textsuperscript{\hspace{0.3em}\raisebox{0.5ex}{\colorbox{green!10}{\tiny\textbf{$\uparrow$1.1}}}} & \textbf{35.8}\textsuperscript{\hspace{0.3em}\raisebox{0.5ex}{\colorbox{green!10}{\tiny\textbf{$\uparrow$2.1}}}} & \textbf{95.3}\textsuperscript{\hspace{0.3em}\raisebox{0.5ex}{\colorbox{green!10}{\tiny\textbf{$\uparrow$0.7}}}} & \textbf{76.3}\textsuperscript{\hspace{0.3em}\raisebox{0.5ex}{\colorbox{green!10}{\tiny\textbf{$\uparrow$0.3}}}} & \textbf{35.8}\textsuperscript{\hspace{0.3em}\raisebox{0.5ex}{\colorbox{green!10}{\tiny\textbf{$\uparrow$8.1}}}} & \textbf{35.6}\textsuperscript{\hspace{0.3em}\raisebox{0.5ex}{\colorbox{green!10}{\tiny\textbf{$\uparrow$8.1}}}} & \textbf{36.1}\textsuperscript{\hspace{0.3em}\raisebox{0.5ex}{\colorbox{green!10}{\tiny\textbf{$\uparrow$8.9}}}} & \textbf{54.9}\textsuperscript{\hspace{0.3em}\raisebox{0.5ex}{\colorbox{green!10}{\tiny\textbf{$\uparrow$4.1}}}} \\
    \bottomrule
    \end{tabular}
    }
    \caption{Downstream tasks results for different settings after training \textbf{3B} models on \textbf{1T} tokens.}
    \label{main results2}
\end{table*}
Table \ref{main results1} and \ref{main results2} present our primary experimental results, revealing several key insights:

\paragraph{Effectiveness of our PDPC framework.} PDPC with \( n=2 \) consistently outperforms all baselines, regardless of the metric used, showing significant performance and convergence improvements over the baseline. Notably, the 3B model trained on 1T tokens with PDPC demonstrates an average accuracy increase of \textbf{4.1\%} across all benchmarks and \textbf{8.1\%} across MMLU and CMMLU, highlighting the effectiveness of our framework. Figure \ref{fig:combinedplot} depicts performance improvements in the 1.3B and 3B models as training progresses, with our method significantly outperforming \textit{Random} in the latter half of pretraining. In this phase, data dominated by high PD is crucial, especially in the 3B model, highlighting the effectiveness of transitioning from low to high PD data, which significantly boosts model performance and promotes emergent capabilities.

\paragraph{The \textit{Sequential} method can somewhat constrain model performance.} Sorting pretraining data by PD from low to high ${\textit{Sequential-PD-Low2High}}$ can outperform \textit{Random} but falls short of the \({\textit{PDPC-PD-S.}}\) strategy. This limitation arises because sorting strictly by PD reduces data diversity, leading to homogeneity that restricts the model's ability to handle complex tasks. In extreme cases, if the dataset contains duplicate samples, a complete sort would likely place identical samples in the same batch, which is detrimental to improving pretraining efficiency.

\paragraph{PD performs better than other metrics in the Preference CL framework. }
Comparison of the results from \textit{Preference CL-PPL-S.}, \textit{Preference CL-Qu.Edu-S.}, and \textit{PDPC-PD-S.} shows that using PD as a metric yields the best results, particularly on the ARC-C and SciQ datasets. PD can accurately reflect the relative difficulty and complexity of samples, which aligns well with the CL principle discussed in Section \ref{sec:intro}. In contrast, relying solely on PPL or educational value may not effectively capture the differences in sample difficulty required for this CL principle.

\subsection{Ablation Study}
\begin{figure}[ht]
    \centering    \includegraphics[width=0.8\linewidth]{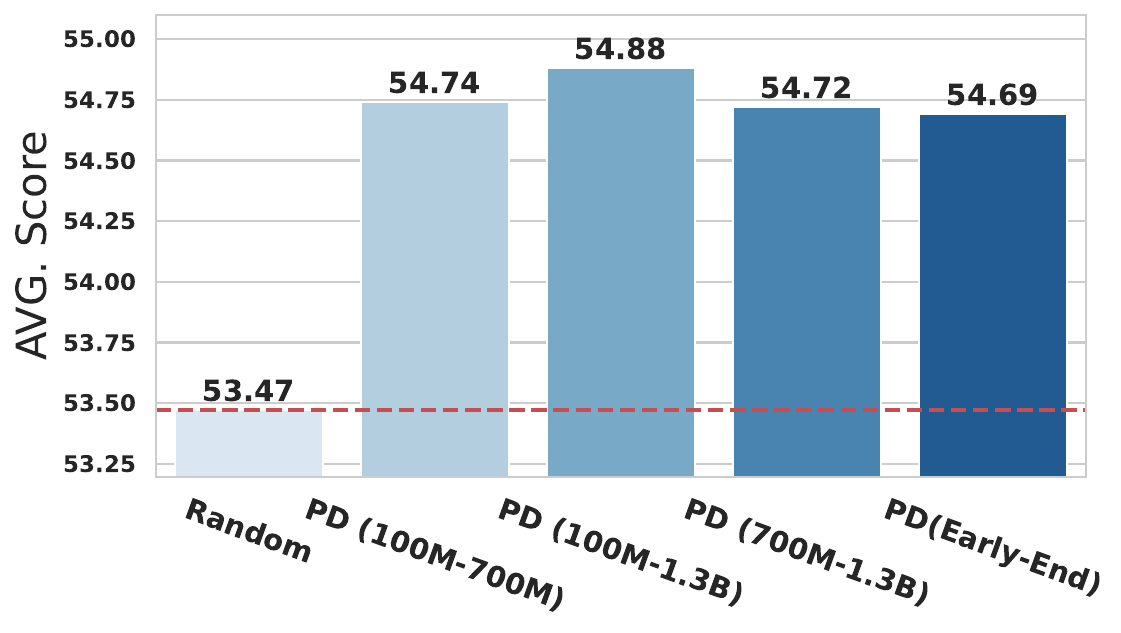}
    \caption{Ablation on PD calculation methods.}
    \label{PD_calculation_methods}
\end{figure}

\paragraph{Impact of PD calculation methods}
We focus on two main factors: \textbf{(1) Size of the RM}: We use PD calculated from various model combinations. \textbf{(2) Choice of the RM}: We use early and late checkpoints from a randomly trained 1.3B model to calculate PD and compare it with PD from RMs.  The experimental results, as shown in Figure \ref{PD_calculation_methods}, demonstrate that regardless of the scale of the RM or the calculation method chosen, the results consistently outperform \textit{Random}. This validates the robustness of our framework regarding PD calculation methods. Additionally, PD from the 100M-700M RMs slightly outperforms that from the early/end models, further supporting our hypothesis that approximating the early and late checkpoints of the model with RMs is valid, as they are comparable in terms of training FLOPS.

\paragraph{Investigation of different preference functions}
Table \ref{function_search} presents the effects of the three preference functions in Section \ref{sec:2.4}. Experimental results indicate that the S-shape function outperforms other functions. Its slower initial decline compared to the linear function highlights the importance of starting with enough low-PD data and gradually introducing high-PD data to enhance performance. The Z-shape function, which uses only low-PD data initially and high-PD data later, slightly outperforms \textit{Random}. The linear, Z-shape, and all S-shape parameter settings outperform \textit{Random}, confirming the robustness of our framework.

\begin{table*}[t]
    \centering
    \setlength{\fboxsep}{1pt} 
    \scalebox{0.9}{
    \small
    \begin{tabular}{c
    >{\centering\arraybackslash}m{2.0cm}| 
    >{\centering\arraybackslash}m{1.4cm}
    >{\centering\arraybackslash}m{1.4cm} 
    >{\centering\arraybackslash}m{1.0cm}
    >{\centering\arraybackslash}m{1.2cm}
    >{\centering\arraybackslash}m{1cm}
    >{\centering\arraybackslash}m{1cm}
    }
    \toprule
        \textbf{Function type} & \textbf{H.P.} & \textbf{ARC-E} & \textbf{ARC-C} & \textbf{SciQ} & \textbf{HellaSw.} & \textbf{PIQA} & \textbf{AVG.} \\ 
    \midrule
        \multicolumn{1}{c}{Random} & - & 56.5 & 23.6 & 85.8 & 34.2 & 67.3 & 53.5 \\
        \multicolumn{1}{c}{Z-shape} & - & 55.6\textsuperscript{\hspace{0.3em}\raisebox{0.5ex}{\colorbox{red!10}{\tiny$\downarrow$0.9}}} & 23.6\textsuperscript{\hspace{0.3em}\raisebox{0.5ex}{{\tiny ~~~0.0}}} & 86.2\textsuperscript{\hspace{0.3em}\raisebox{0.5ex}{\colorbox{green!10}{\tiny$\uparrow$0.4}}} & 33.8\textsuperscript{\hspace{0.3em}\raisebox{0.5ex}{\colorbox{red!10}{\tiny$\downarrow$0.4}}} & \textbf{68.9}\textsuperscript{\hspace{0.3em}\raisebox{0.5ex}{\colorbox{green!10}{\tiny$\uparrow$1.6}}} & 53.6\textsuperscript{\hspace{0.3em}\raisebox{0.5ex}{\colorbox{green!10}{\tiny$\uparrow$0.1}}} \\ 
        \multicolumn{1}{c}{Linear} & - & 55.5\textsuperscript{\hspace{0.3em}\raisebox{0.5ex}{\colorbox{red!10}{\tiny$\downarrow$1.0}}} & 23.9\textsuperscript{\hspace{0.3em}\raisebox{0.5ex}{\colorbox{green!10}{\tiny$\uparrow$0.3}}} & 87.7\textsuperscript{\hspace{0.3em}\raisebox{0.5ex}{\colorbox{green!10}{\tiny$\uparrow$1.9}}} & 34.2\textsuperscript{\hspace{0.3em}\raisebox{0.5ex}{{\tiny ~~~0.0}}} & 67.3\textsuperscript{\hspace{0.3em}\raisebox{0.5ex}{{\tiny ~~~0.0}}} & 53.7\textsuperscript{\hspace{0.3em}\raisebox{0.5ex}{\colorbox{green!10}{\tiny$\uparrow$0.2}}} \\
    \midrule
        \multirow{4}{*}{S-shape} & a=2.5 & 57.1\textsuperscript{\hspace{0.3em}\raisebox{0.5ex}{\colorbox{green!10}{\tiny$\uparrow$0.6}}} & 24.8\textsuperscript{\hspace{0.3em}\raisebox{0.5ex}{\colorbox{green!10}{\tiny$\uparrow$1.2}}} & 87.6\textsuperscript{\hspace{0.3em}\raisebox{0.5ex}{\colorbox{green!10}{\tiny$\uparrow$1.8}}} & 34.0\textsuperscript{\hspace{0.3em}\raisebox{0.5ex}{\colorbox{red!10}{\tiny$\downarrow$0.2}}} & 67.5\textsuperscript{\hspace{0.3em}\raisebox{0.5ex}{\colorbox{green!10}{\tiny$\uparrow$0.2}}} & 54.2\textsuperscript{\hspace{0.3em}\raisebox{0.5ex}{\colorbox{green!10}{\tiny$\uparrow$0.7}}} \\ 
        ~ & a=5.0 & \textbf{57.5}\textsuperscript{\hspace{0.3em}\raisebox{0.5ex}{\colorbox{green!10}{\tiny$\uparrow$1.0}}} & 26.0\textsuperscript{\hspace{0.3em}\raisebox{0.5ex}{\colorbox{green!10}{\tiny$\uparrow$2.4}}} & 87.7\textsuperscript{\hspace{0.3em}\raisebox{0.5ex}{\colorbox{green!10}{\tiny$\uparrow$1.9}}} & \textbf{34.3}\textsuperscript{\hspace{0.3em}\raisebox{0.5ex}{\colorbox{green!10}{\tiny$\uparrow$0.1}}} & 67.4\textsuperscript{\hspace{0.3em}\raisebox{0.5ex}{\colorbox{red!10}{\tiny$\downarrow$0.1}}} & 54.6\textsuperscript{\hspace{0.3em}\raisebox{0.5ex}{\colorbox{green!10}{\tiny$\uparrow$1.1}}} \\  
        ~ & a=7.5 & 56.4\textsuperscript{\hspace{0.3em}\raisebox{0.5ex}{\colorbox{red!10}{\tiny$\downarrow$0.1}}} & 25.4\textsuperscript{\hspace{0.3em}\raisebox{0.5ex}{\colorbox{green!10}{\tiny$\uparrow$1.8}}} & 87.2\textsuperscript{\hspace{0.3em}\raisebox{0.5ex}{\colorbox{green!10}{\tiny$\uparrow$1.4}}} & 34.2\textsuperscript{\hspace{0.3em}\raisebox{0.5ex}{{\tiny ~~~0.0}}} & \textbf{68.9}\textsuperscript{\hspace{0.3em}\raisebox{0.5ex}{\colorbox{green!10}{\tiny$\uparrow$1.6}}} & 54.4\textsuperscript{\hspace{0.3em}\raisebox{0.5ex}{\colorbox{green!10}{\tiny$\uparrow$0.9}}} \\
        ~ & a=10.0 & 57.3\textsuperscript{\hspace{0.3em}\raisebox{0.5ex}{\colorbox{green!10}{\tiny$\uparrow$0.8}}} & \textbf{26.6}\textsuperscript{\hspace{0.3em}\raisebox{0.5ex}{\colorbox{green!10}{\tiny$\uparrow$3.0}}} & \textbf{87.9}\textsuperscript{\hspace{0.3em}\raisebox{0.5ex}{\colorbox{green!10}{\tiny$\uparrow$2.1}}} & 33.7\textsuperscript{\hspace{0.3em}\raisebox{0.5ex}{\colorbox{red!10}{\tiny$\downarrow$0.5}}} & 68.0\textsuperscript{\hspace{0.3em}\raisebox{0.5ex}{\colorbox{green!10}{\tiny$\uparrow$0.7}}} & \textbf{54.7}\textsuperscript{\hspace{0.3em}\raisebox{0.5ex}{\colorbox{green!10}{\tiny\textbf{$\uparrow$1.2}}}} \\
    \bottomrule
    \end{tabular}
    }
    \caption{Downstream tasks results for different preference functions. We report accuracy for each task, and the best performances are marked in bold. Abbreviations: H.P. = Hyper-parameters.}
    \label{function_search}
\end{table*}
\subsection{Analysis}
\begin{figure}[t]
    \centering
    \begin{subfigure}{0.48\linewidth}
        \centering
        \includegraphics[width=\linewidth]{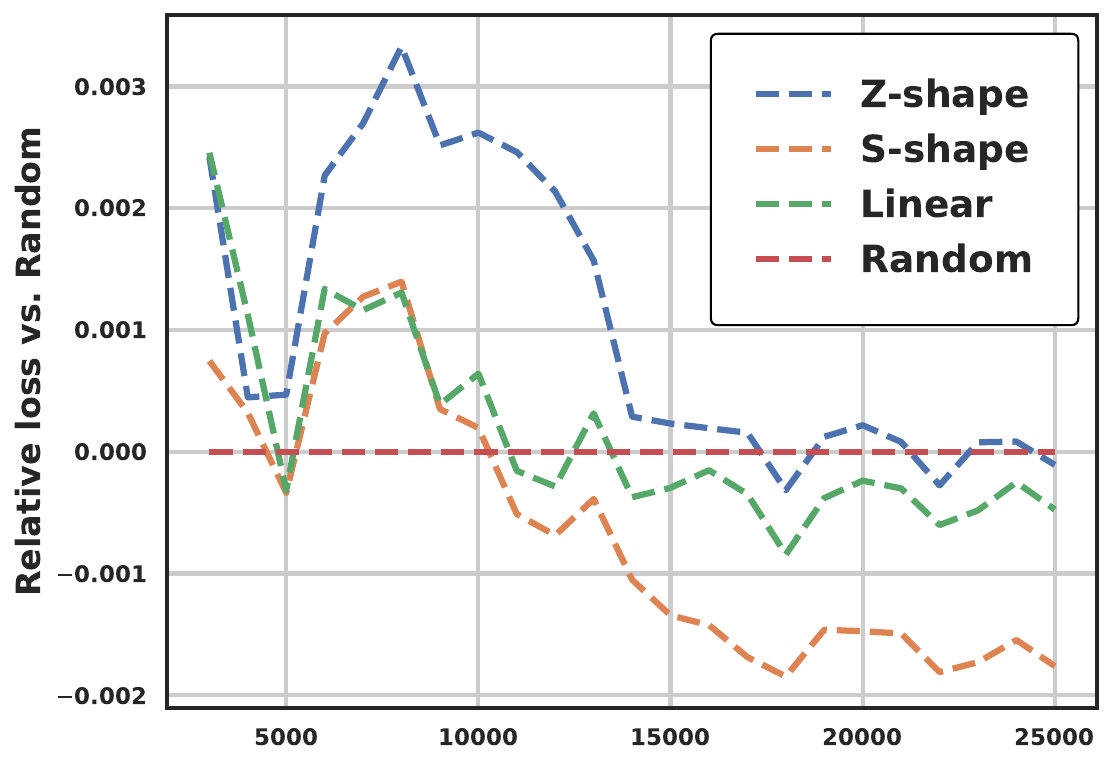}
        \caption{Relative loss vs. Random.}
        \label{test_loss}
    \end{subfigure}
    \hfill
    \begin{subfigure}{0.48\linewidth}
        \centering
        \includegraphics[width=\linewidth]{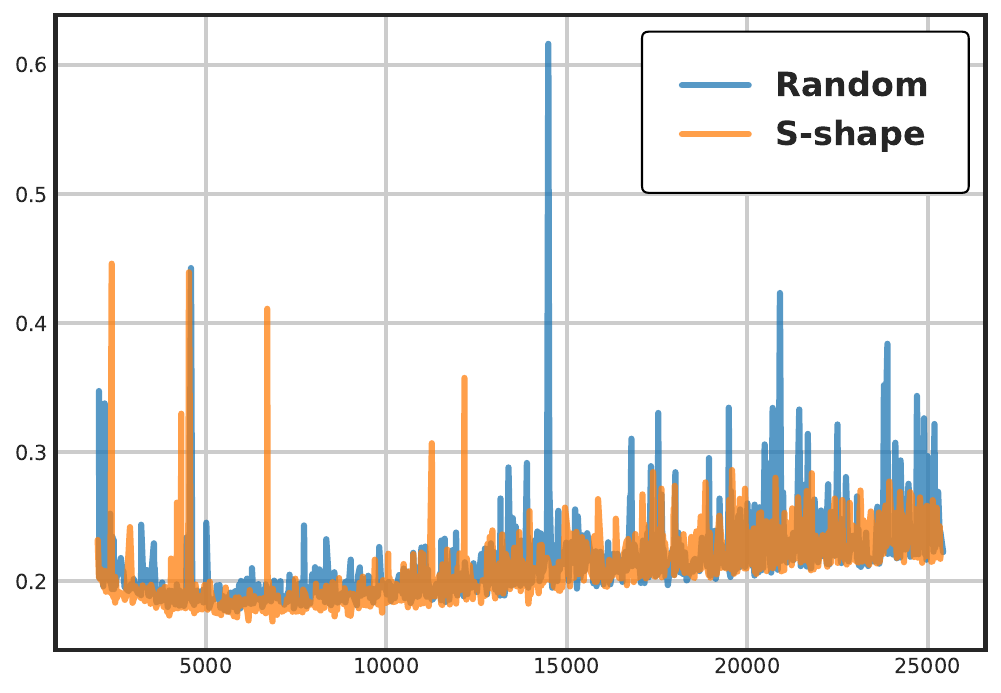}
        \caption{1B gradient norm w/ steps.}
        \label{grad_norm}
    \end{subfigure}

    \begin{subfigure}{0.48\linewidth}
        \centering
        \includegraphics[width=\linewidth]{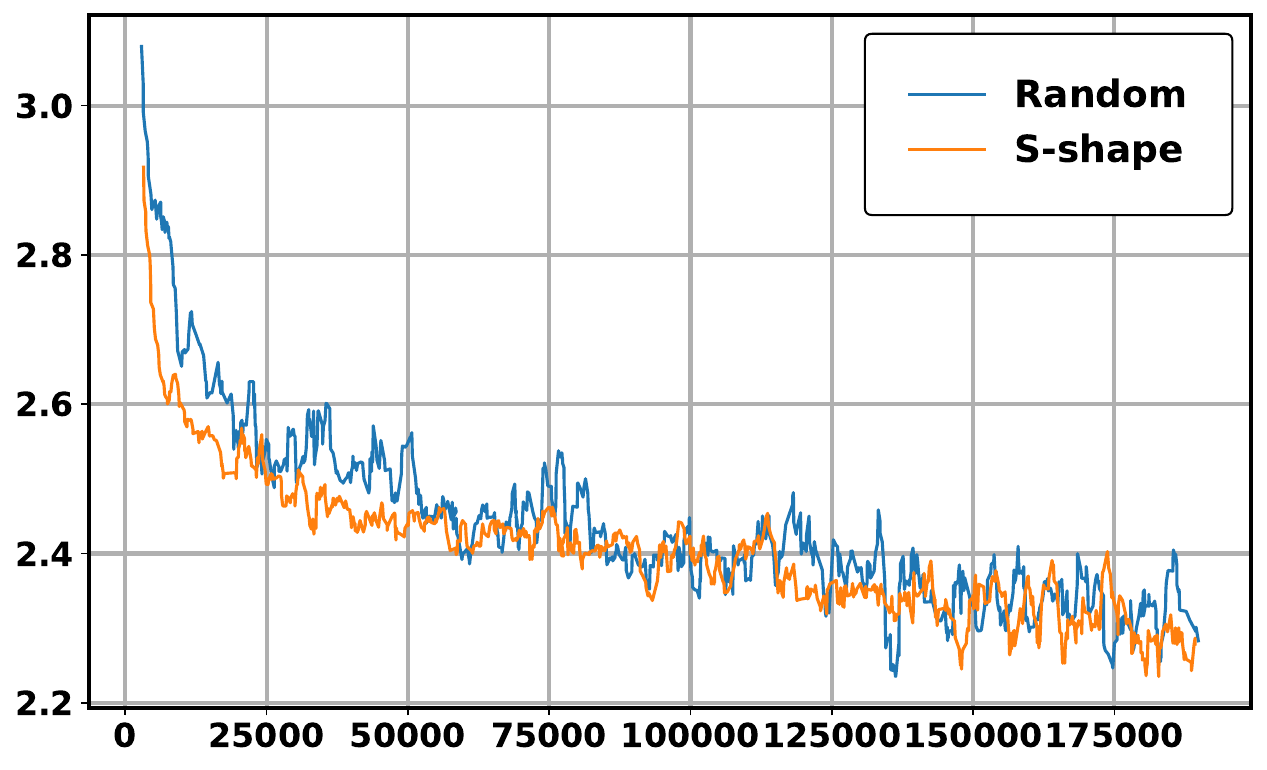}
        \caption{Training loss vs. Random.}
        \label{3B_train_loss}
    \end{subfigure}
    \hfill
    \begin{subfigure}{0.48\linewidth}
        \centering
        \includegraphics[width=\linewidth]{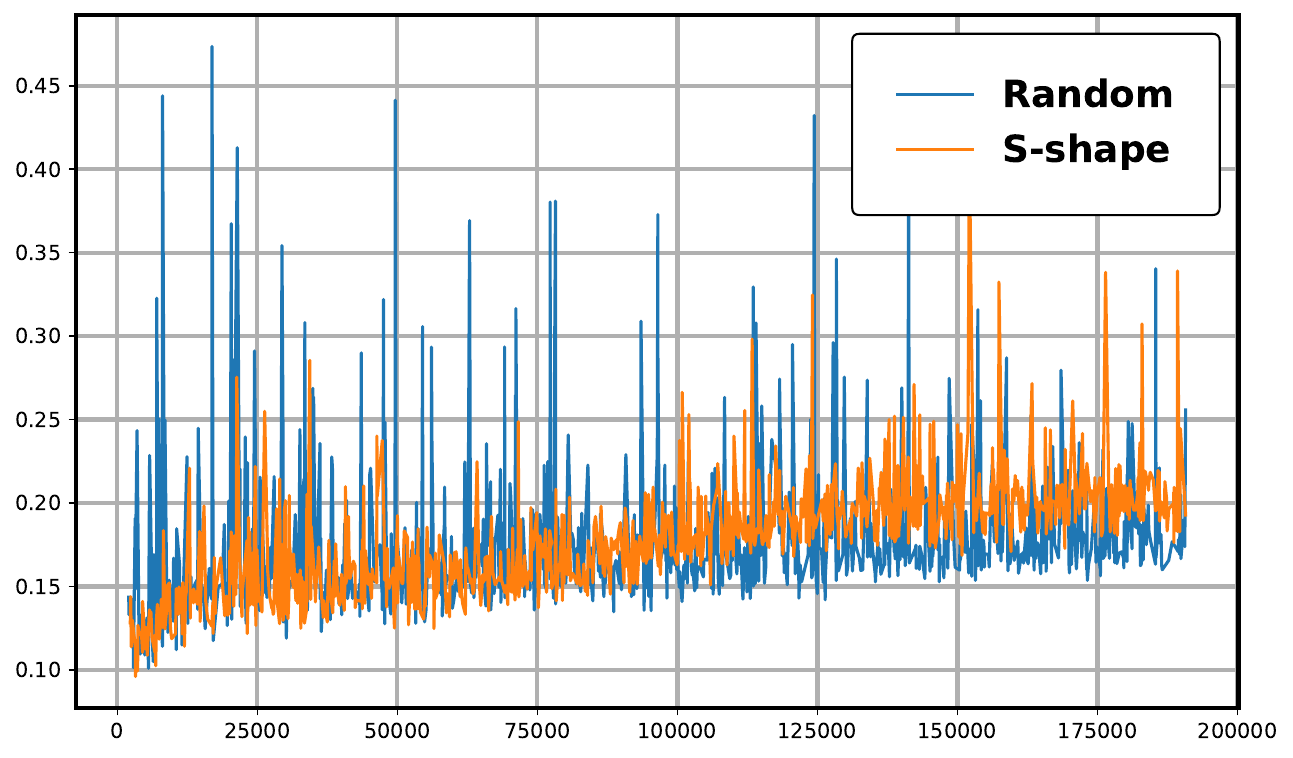}
        \caption{3B gradient norm w/ steps.}
        \label{3B_grad_norm}
    \end{subfigure}
    \caption{(a) Relative test loss and (b) gradient norm during model training of 1B model. (c)Training loss and (d) gradient norm during 3B model training.}
    \label{combined_3B_figures}
\end{figure}

\paragraph{Loss Analysis} We sample 500M tokens from SlimPajama as test set to compare test loss with $Random$ on the 1.3B setting. Figure \ref{combined_3B_figures}(a) shows that PDPC's test loss initially declines slowly, then rapidly decreases, achieving a lower loss than $Random$ by incorporating higher PD data later in training. The S-shape function effectively helps to minimize loss. Additionally, Figure \ref{combined_3B_figures}(b) demonstrates that our method stabilizes the gradient norm, ensuring smoother model convergence. In the 3B model experiments (Figure \ref{combined_3B_figures}(c) and \ref{combined_3B_figures}(d)), the S-shape function significantly outperforms $Random$. It accelerates early loss reduction, speeds up convergence, and achieves a lower final loss.

\subsection{Case study} \label{sec:case}
\begin{figure}[htp]
        \centering 
        \begin{subfigure}{0.49\linewidth}
        \centering        \includegraphics[width=\linewidth]{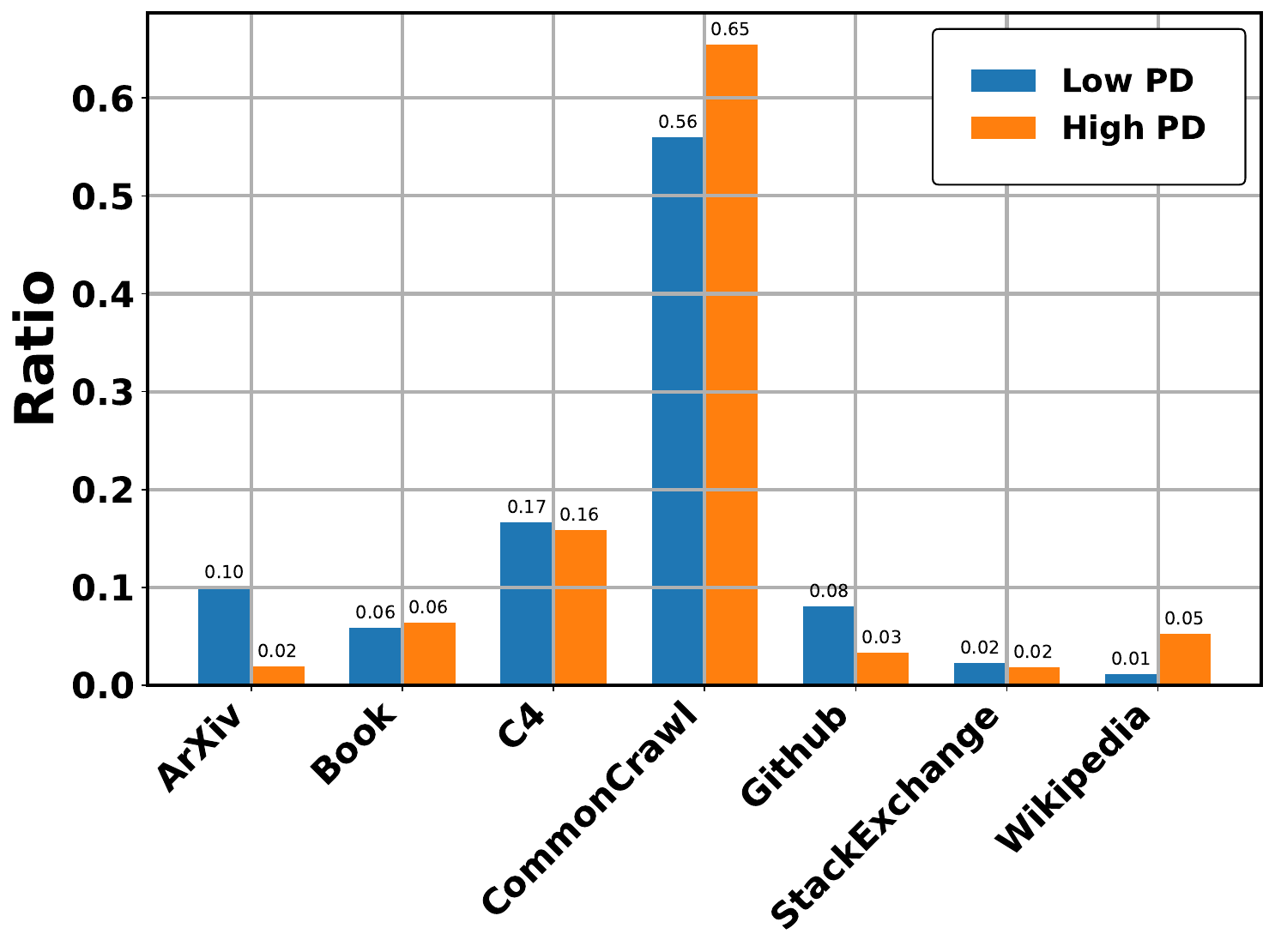}
        \caption{Normalization along domains.}
        \label{pdd}
        \end{subfigure}
        \hfill
    \begin{subfigure}{0.49\linewidth}
        \centering
        \includegraphics[width=\linewidth]{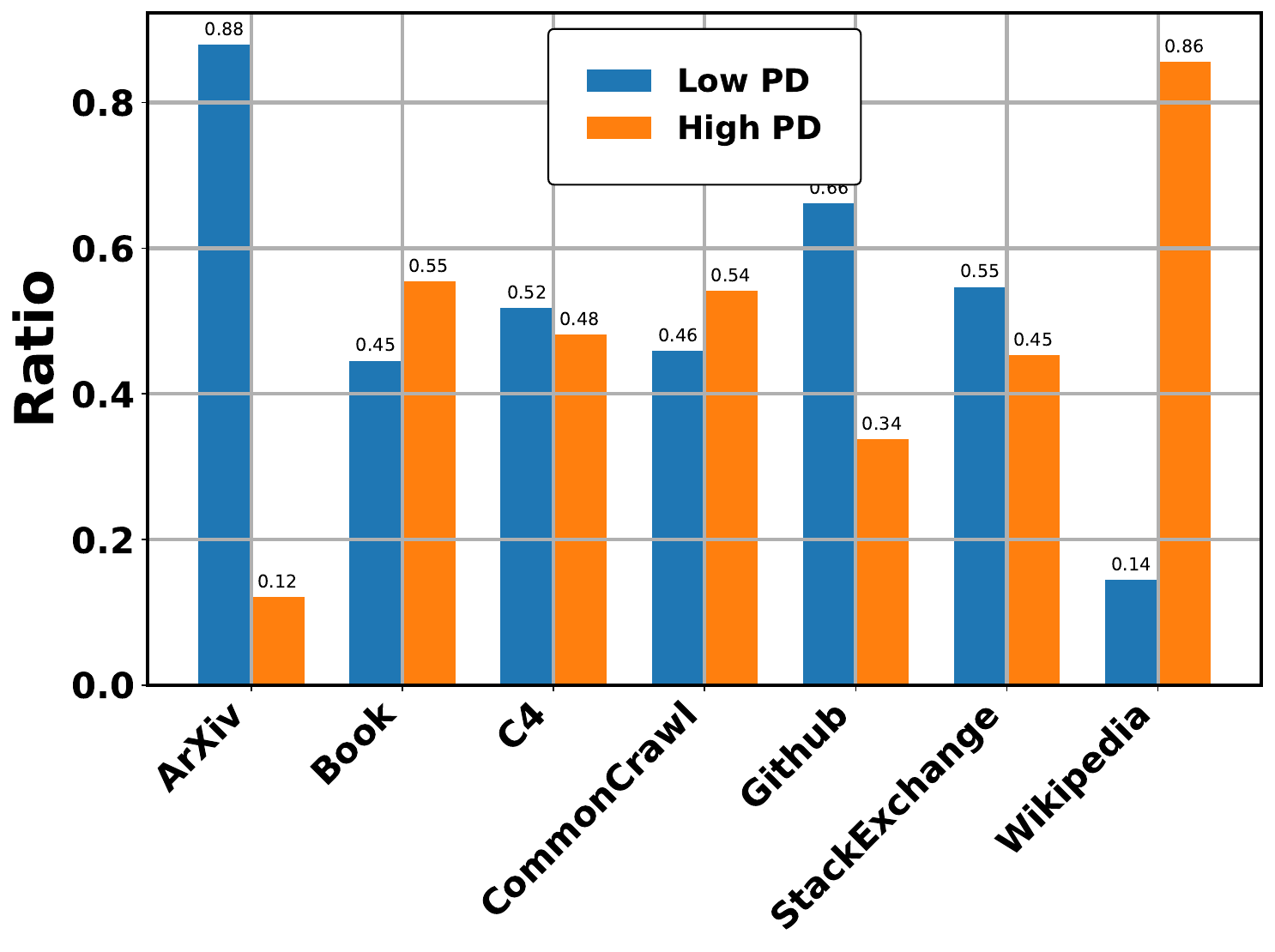}
        \caption{Normalization along PD partitions.}
        \label{pdp}
    \end{subfigure}
    \caption{Data distribution across different sources.}
     \label{data_distribution}
\end{figure}

\paragraph{Data Source Distribution} Our analysis of 25M texts from Slimpajama \cite{cerebras2023slimpajama} reveals significant differences in distribution and semantics between high-PD and low-PD data. High-PD data mainly comes from Wikipedia and CommonCrawl, while low-PD data is sourced from arXiv and GitHub, as seen in Figure \ref{data_distribution}. The significant performance improvement of PDPC in later stages is largely attributed to the higher proportion of high-quality data from sources like Wikipedia, which primarily appears in high-PD data and is less prevalent in low-PD data. 

\begin{figure}[t]
    \centering
    \begin{subfigure}{0.44\linewidth}
        \centering
        \includegraphics[width=\linewidth]{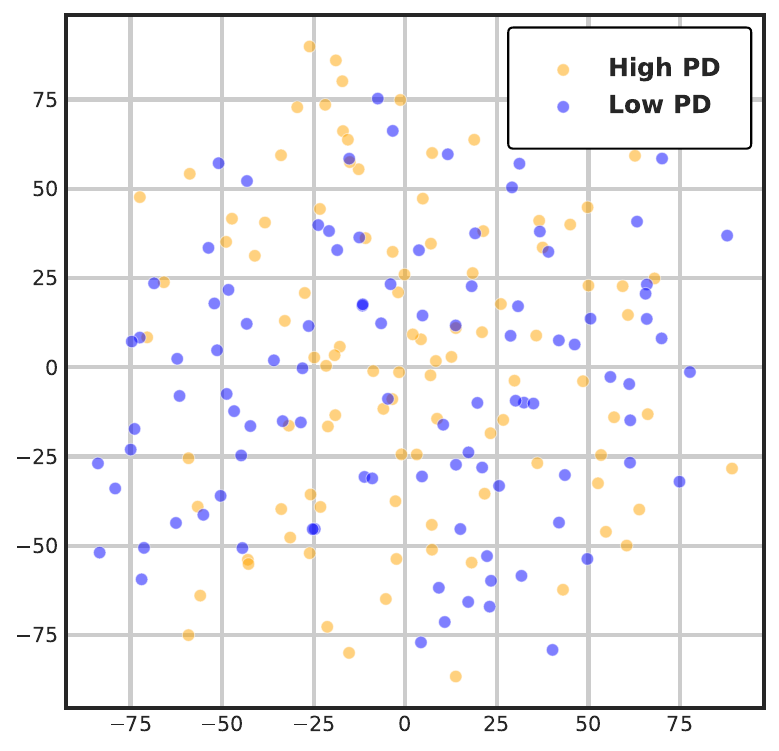}
        \caption{Uniform sampling.}
        \label{diversity_uni}
    \end{subfigure}
    \hfill
    \begin{subfigure}{0.44\linewidth}
        \centering
        \includegraphics[width=\linewidth]{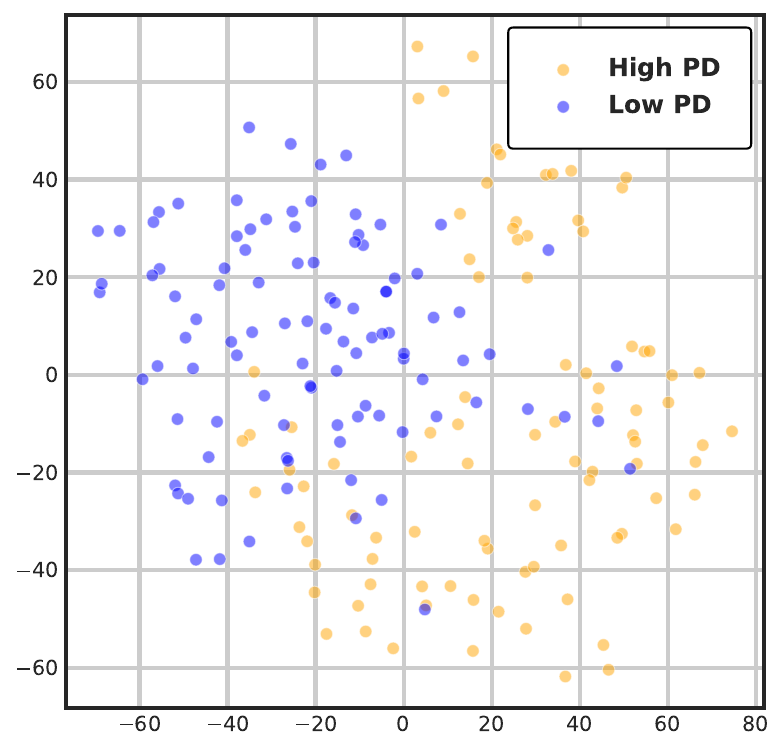}
        \caption{Extreme sampling.}
        \label{diversity}
    \end{subfigure}
    \caption{Analysis of semantic distributions.}
    \label{combined_figures2}
\end{figure}

To understand the semantic structure of the data, we use T5 \cite{raffel2023exploringlimitstransferlearning} to generate dense vector representations of texts collected in two ways: (a) uniformly from different PD partitions, and (b) from extreme PD intervals (top/bottom 10\%) after sorting. We then apply t-SNE for dimensionality reduction. Figure \ref{combined_figures2} illustrates the semantic visualization of the data points. Uniform sampling results in high and low PD data being evenly distributed in semantic space, indicating semantic diversity. In contrast, extreme PD sampling leads to distinct semantic spaces, explaining the suboptimality of \textit{Sequential-PD-Low2High}, as it may result in a lack of data diversity in some batches during training, thereby affecting model performance.

\begin{figure}[t]
    \centering
    \includegraphics[width=0.75\linewidth]{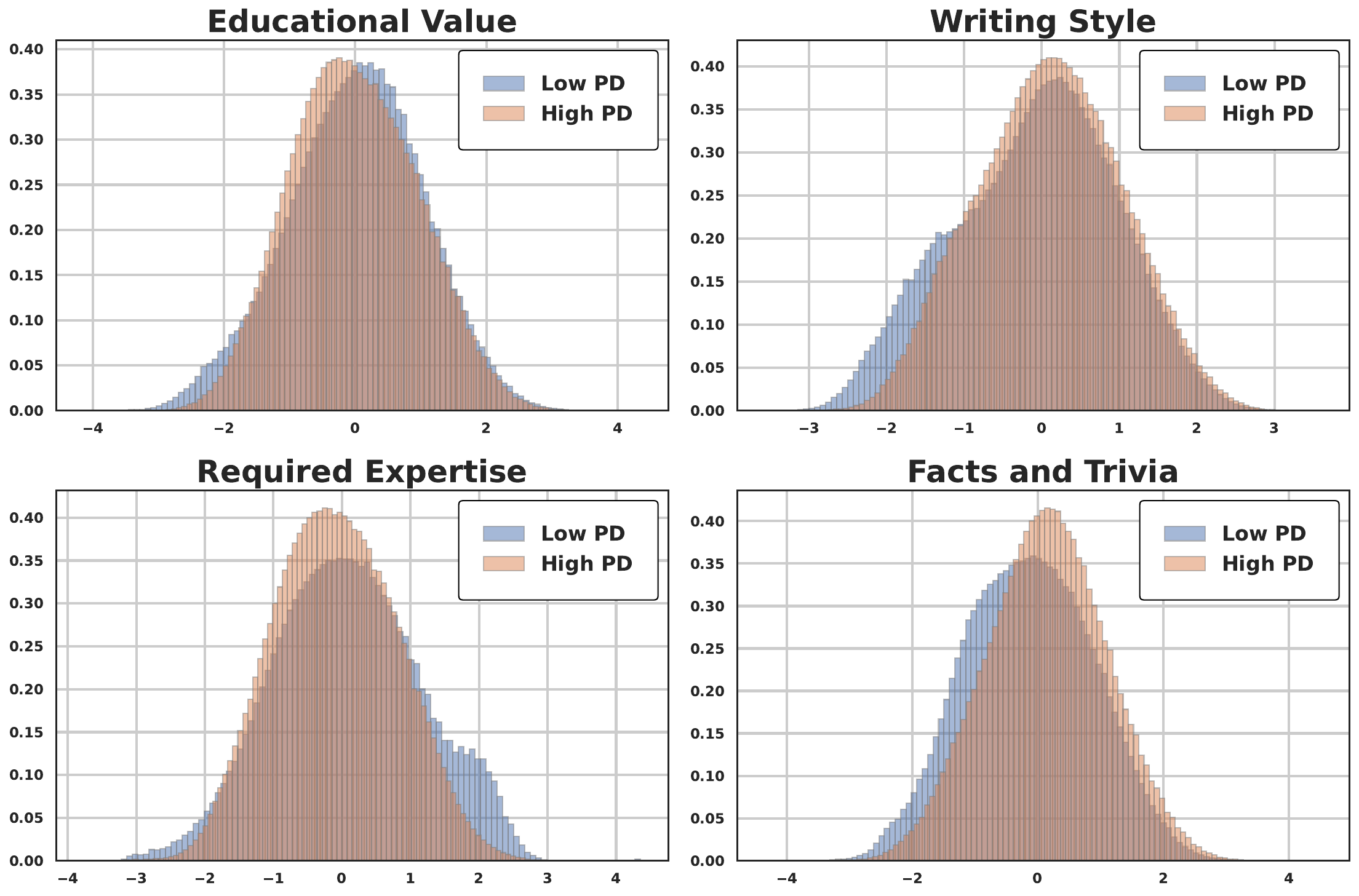}
    \caption{Distribution of low-PD and high-PD data across different quality dimensions.}
    \label{quality_distribution}
\end{figure}

\paragraph{Data Quality Distribution} We use 4 raters from QuRating\cite{wettig2024qurating} to assess data quality. Figure \ref{quality_distribution} shows consistent quality distributions in both low-PD and high-PD parts, ensuring uniform quality throughout the pretraining process and preventing the model from learning from lower-quality data at any stage.
\paragraph{Stability of PD} We evaluate the Spearman correlation coefficients between different PD types (detailed in Figure \ref{corr_heat_map} of Appendix \ref{c3}) and find a strong correlation among PDs derived from RMs of varying sizes, which indicates that PD is a relatively stable metric. Calculating PD with smaller RMs yields results consistent with larger RMs, saving computational resources. Furthermore, larger size discrepancies among RMs result in broader PD distributions, which enhance data differentiation (detailed in Figure \ref{PD__distribution} of Appendix \ref{c2}). This finding is supported by ablation tests, which show that PD calculations using models ranging from 100M to 1.3B yield the best results. Additionally, PD maintains a consistent distribution across domains. For instance, the PD between 100M and 700M models generally appears to follow a normal distribution with an approximate mean of 0.27. Partitioning and organizing data using PD ensures that the data at each training step does not skew towards specific sub-domains, allowing the model to encounter a diverse range of data throughout the entire training process.

\paragraph{Semantic Properties Analysis} To explore the semantic features of high-PD and low-PD data, we analyze 1,000 samples randomly sampled from each part using 10 criteria focused on semantic features. These criteria encompass polysemy, specialized terminology, cultural context, logical reasoning, humor, ethical dimensions, intricate sentence structures, scientific concepts, emotional nuances, and background knowledge. Each criterion is clearly defined for GPT-4o to assess with "yes" or "no" responses. More details about the prompt can be found in the Appendix \ref{sec:prompt}. Figure \ref{fig:properties} shows that PD is independent of other linguistic features. Partitioning and organizing data with PD maintains diversity in semantic properties, ensuring that model performance is not restricted by data homogeneity.
\begin{figure}[t]
    \centering    
    \includegraphics[width=0.8\linewidth]{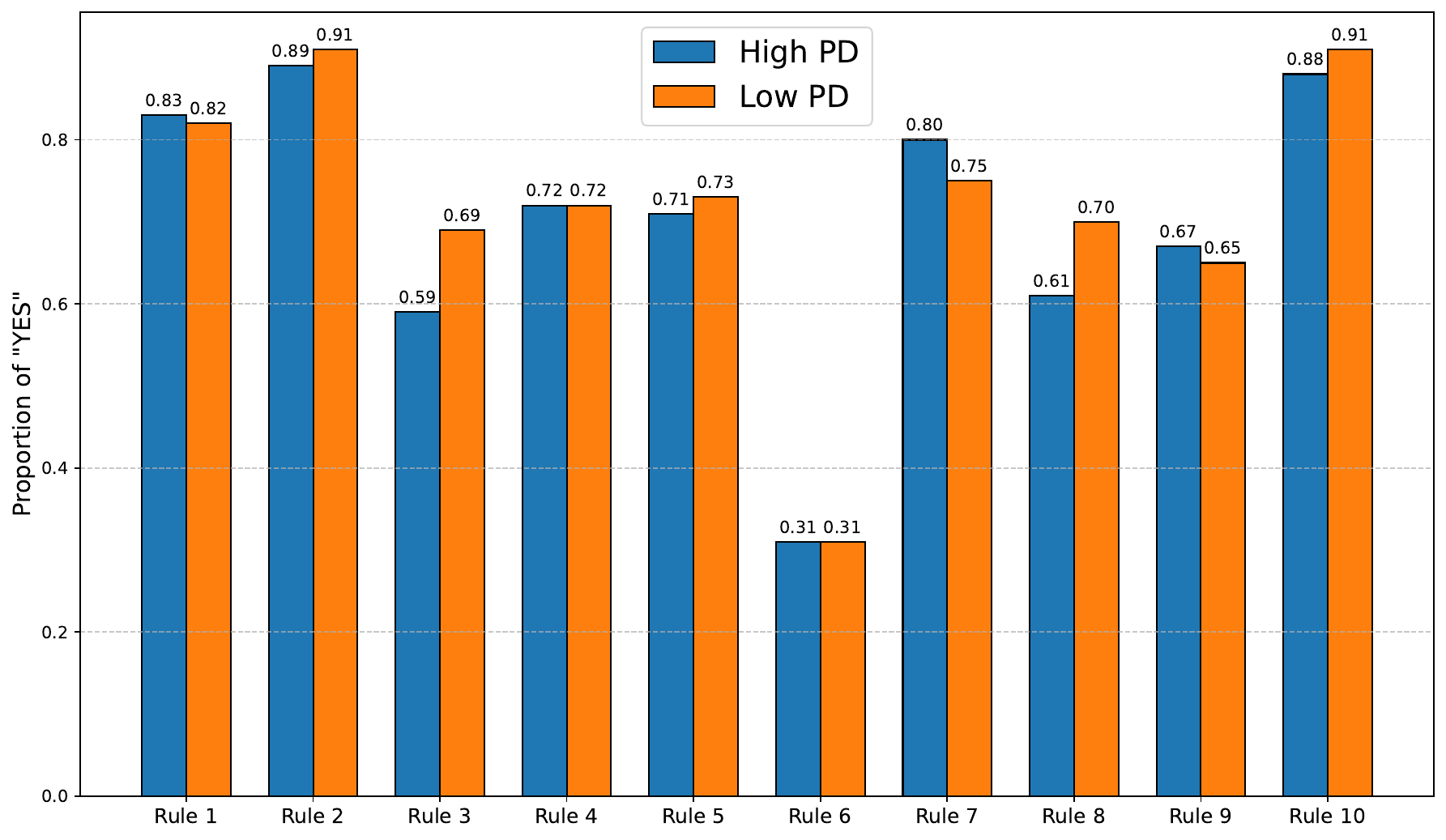}
    \caption{Semantic properties differences of low-PD data and high-PD data.}
    \label{fig:properties}
\end{figure}
\vspace{-10mm}
\section{Related Works}
Data preprocessing is crucial in LLM pretraining, ensuring dataset quality and integrity. Traditional methods use expert-crafted rules to filter low-quality data and remove duplicates \cite{raffel2020exploring, rae2021scaling, laurenccon2022bigscience, together2023redpajama, penedo2024fineweb}. Enhanced approaches leverage target data sources or proxy models for curation \cite{wenzek2020ccnet, xie2023data, marion2023less}. Automated data selection using classifiers is gaining traction; for example, \citeauthor{du2022glam} (\citeyear{du2022glam}) employed logistic regression to evaluate data quality, and other studies have developed sophisticated scoring mechanisms \cite{zhang2024autonomous, sachdeva2024train}. QuRating \cite{wettig2024qurating} uses multiple raters to assess data contributions. Curriculum Learning (CL) complements these efforts by organizing training data from simple to complex, improving learning efficiency and generalization \cite{forestier2022intrinsicallymotivatedgoalexploration, soviany2021curriculumselfpacedlearningcrossdomain}. In NLP, CL enhances models, such as in word embeddings \cite{collobert2008unified} and neural machine translation \cite{platanios2019competence}. Recently, CL's application in LLM pretraining is also growing \cite{wu2024curriculum}.

\section{Conclusion}
In this paper, we propose PDPC to address the limitations of consistent data distribution in pretraining LLMs. PDPC perceives the models' preferences and utilizes different, model-preferred data as the models' capabilities improve to enhance their performance. We introduce PD as a data metric and incorporate the preference function based on PD to predict data preferences, enabling the offline organization of data and ensuring uninterrupted pretraining. Experiments show that PDPC significantly outperforms the baselines, with the 3B model achieving an average improvement of \textbf{8.1}\% over \textit{Random} on MMLU and CMMLU. 

\section{Limitations and Future Works}
\paragraph{Exploration of additional PD partitions}  
This study primarily focuses on the scenario where \( n = 2 \), analyzing concentration mixing curves and systematically blending two subsets with higher and lower PD in accordance with training progression. However, we have not yet explored dividing the training data into more than two subsets to assess whether further performance enhancements are attainable. In future research, we plan to investigate cases where \( n > 2 \) and develop novel methodologies for addressing learning curves.

\paragraph{Iterative update of learning curves}  
We determine the S-shaped learning curve through functional exploration and use it as the basis for arranging the data sequence to train the model. In fact, we can also start from the newly trained model, re-explore new learning curves, and iteratively update our curriculum learning path. Optimizing the learning curve through multiple iterations could be one of our future research directions.


\bibliography{main}

\appendix

\section{Ethical Considerations}
Due to the influence of training data, LLMs are prone to generating untruthful or socially harmful content. We aim to mitigate this issue by enhancing the reliability of model training and the model’s final performance through the proposed training data adjustment framework. Additionally, training LLMs incurs substantial time and financial costs. Therefore, exploring ways to maximize the efficiency of training data utilization will be key to addressing this problem and can also contribute to reducing global carbon emissions.

\section{Preliminary Exploration of Iterative Optimization of Preference Functions}
\label{sec:apc}
When employing a grid search methodology, the size of the solution space scales as \( n^T \), where \( T \) represents the number of training steps. Consequently, an increase in the number of parts \( n \) results in an exponential expansion of the solution space.

In Section \ref{sec:2.4}, we introduce a curriculum learning method that partitions pretraining data into 2 parts and identifies the S-shape preference function through theoretical analysis. This method is simple and efficient. However, in resource-rich scenarios, we also offer a more precise approach to simulate the model's preferences at different pretraining stages, as shown in Algorithm \ref{alg:2}. Specifically, after using the discovered preference function to guide the model's pretraining, we conduct annealing experiments on checkpoints from different stages to explore the model's preferences for data mixing ratios. Based on these preferences, we fit the model's preference function to guide the next round of pretraining. This process is iterated until the model's performance converges.

\subsection{Proportion Preference Annealing Experiment}
In this subsection, we aim to systematically explore the model's preference for data mixing ratios based on PD at different pretraining stages. This process is crucial for understanding the dynamic changes in model preferences during the pretraining process and provides empirical evidence for optimizing curriculum learning strategies.

Firstly, we construct the dataset required for the annealing experiment. Based on the median of PD values, samples are divided into two parts: low-PD data and high-PD data. We create 11 different annealing datasets where the proportion of low-PD data takes values of 0\%, 10\%, 20\%, ..., 100\%. To enhance data diversity, each dataset is supplemented with 30\% of samples that share the same distribution as the pretraining data, and the mixed samples remain consistent across all datasets. This design ensures the robustness and comparability of the experimental results.

The annealing experiments are conducted at various stages of model pretraining to evaluate the model's preference for data mixing ratios at different training progressions. We perform experiments on checkpoints from 8 pretraining stages, corresponding to pertraining progress of 0\%, 12.5\%, 25\%, 37.5\%, 50\%, 62.5\%, 75\%, 87.5\%, and 100\%. At each checkpoint, we evaluate the model using all 11 annealing datasets and record the model's performance across different mixing ratios.

By conducting annealing experiments at checkpoints throughout the pretraining stages, we obtain a series of preference data points \((p, b)\), where \( p \) represents the pretraining progress, and \( b \) denotes the model's preference for the proportion of low-PD data at that stage. Specifically, for each stage \( p \), \( b \) is defined as the proportion of low-PD data that optimizes model performance, i.e.,
\begin{equation}
    b = \arg\max_{\beta} \mathcal{M}(\beta)
\end{equation}
where \(\mathcal{M}(\beta)\) denotes the comprehensive performance metric of the model on the annealing dataset with the proportion of low-PD data \( \beta \).

In our study, to fit the changes of preference of LLMs during pretraining, we employ the Piecewise Cubic Hermite Interpolating Polynomial (PCHIP) method. Given experimental data points \((p_i, b_i)\), the PCHIP method constructs local cubic polynomials to ensure monotonicity and smoothness within each interval. Specifically, for each interval \([p_i, p_{i+1}]\), PCHIP generates a cubic polynomial:
\begin{equation}
    S_i(p) = a_i(p - p_i)^3 + b_i(p - p_i)^2 + c_i(p - p_i) + d_i
\end{equation}
where coefficients \(a_i, b_i, c_i, d_i\) are determined by satisfying interpolation conditions, derivative continuity, and monotonicity conditions. These conditions ensure that the fitting curve not only passes through all data points but also maintains monotonicity within each interval, preventing overfitting.

We apply the interpolation function to a uniformly distributed set of points from 0 to 1 to obtain a continuous function curve of concentration preference \( b \) as it varies with pretraining progress \( p \). We constrain the values of the fitted curve between 0 and 1. This method effectively captures the trend of model preferences for datasets at different pretraining stages, providing a reliable foundation for subsequent analysis.

\begin{algorithm}[t] 
\caption{\small Iterative Optimization of Preference Functions}
\small
\begin{algorithmic}[1]
\State \textbf{Input:} Pretraining dataset $\mathcal{D}$, initial preference function $f(p)$, termination threshold $\epsilon$
\State \textbf{Output:} Iteratively trained model \(\theta_K\)
\While{not converged}
    \State Partition \( \mathcal{D} \) into \( 2 \) sub-domains \( A_{PD}^{low} \) and $A_{PD}^{high}$. 
    \State Construct the annealing dataset \(\mathcal{D}_i\) with the proportion of low-PD data set to \(\beta_i\): 
    \State $\beta_i \in \{0\%, 10\%, \ldots, 100\%\}$
    \For{$p$ in $\{0\%, 12.5\%, \ldots, 100\%\}$}
        \State Retrieve the model checkpoint \(\theta_p\)
        \State Evaluate model performance $\mathcal{M}_p(c)$ on \(\{\mathcal{D}_i\}_{i=0}^{10}\)
        \State Record preference $b_p = \arg\max_{c} \mathcal{M}_p(c)$
    \EndFor
    \State Use PCHIP to fit $b = f(p)$ from data points $(p_i, b_i)$
    \If{change in $f(p)$ between iterations $< \epsilon$}
        \State Break
    \EndIf
    \For{$k = 0$ \textbf{to} $K-1$}
    \State Calculate pretraining progress  $p = \frac{k}{K}$
    \State Get the proportions vector: 
    \State $[\alpha_{1}, \alpha_{2}] \gets [f(p), 1-f(p)]$
    \State Sample from the two domains to form $B_{k}$:
    \State $B_{k} = \{ x \sim A_{PD}^{low}\}_{\alpha_{1} \cdot N} \cup \{ x \sim A_{PD}^{high}\}_{\alpha_{2} \cdot N}$
    \State Train the model on \( B_k \) and update \(\theta_k\)
    \EndFor
    
\EndWhile
\end{algorithmic}
\label{alg:2}
\end{algorithm}

\subsection{Iterative Curriculum Learning}
After completing the proportion preference annealing experiment and successfully fitting the preference function \( b = f(p) \), we apply this function to optimize curriculum learning strategies, guiding the pretraining process of the model. Specifically, based on the fitted function \( f(p) \), we dynamically determine the optimal proportion \( b \) of low-PD data during training according to the current pretraining completion \( p \). 

However, it is important to note that the integral of the fitted preference function over the interval \([0, 1]\) may not equal 0.5. This implies that, under this configuration, the amounts of low-PD and high-PD data used may not be equal. To ensure the reasonable utilization of all data, we calculate the integral of the function over \([0, 1]\) to determine the quantile threshold for dataset division. The specific formula is:
\begin{equation}
    \int_0^1 f(p) \, dp = \alpha
\end{equation}
where \(\alpha\) guides to adjust the data allocation ratio.

After pretraining is completed, we conduct the proportion preference annealing experiment again to obtain updated preference data points \((p_i, b_i)\), and refit the preference function \( f(p) \) based on these data. This process continues iteratively until the following termination condition is met: the change in the preference function between two consecutive iterations is below a predefined threshold. This method ensures that the model's data preference is precisely adjusted and optimized.

\section{Experimental Details}
\subsection{PPL Distribution of low-PD and high-PD data} 
\begin{figure}[htbp]
    \centering
    \begin{subfigure}[b]{0.7\linewidth}
        \includegraphics[width=\textwidth]{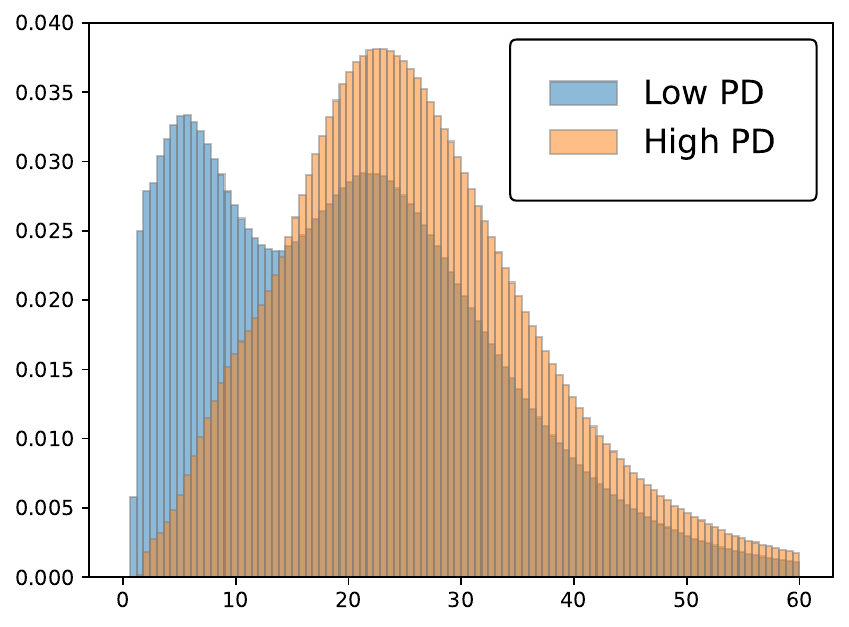}
        \caption{PPL of 100M RM.}
        \label{ppl_low_pd}
    \end{subfigure}
    \hfill
    \begin{subfigure}[b]{0.7\linewidth}
        \includegraphics[width=\textwidth]{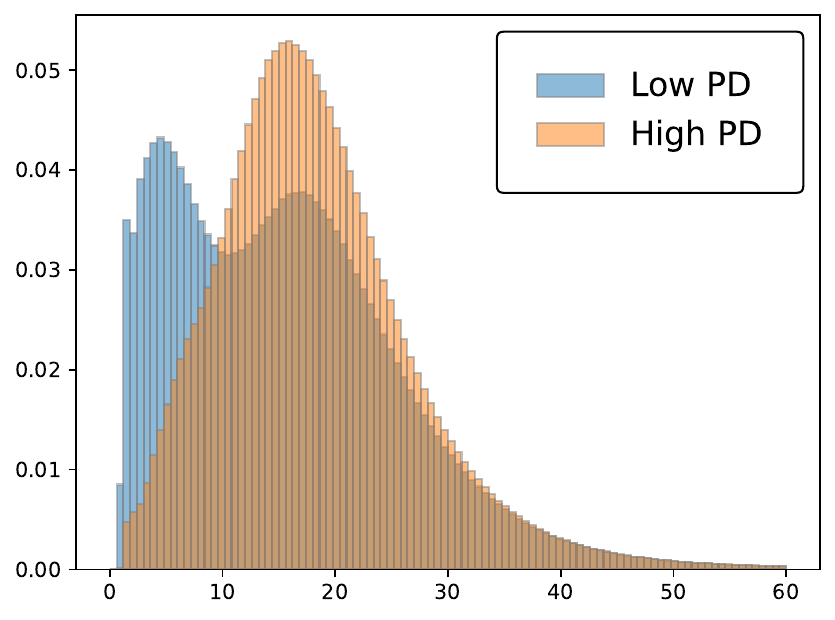}
        \caption{PPL of 700 RM.}
        \label{ppl_high_pd}
    \end{subfigure}
    \caption{PPL distribution of low-PD and high-PD data.}
    \label{ppl_distribution}
\end{figure}

As shown in Figure \ref{ppl_distribution}, examining the perplexity distribution across data with varying PD values reveals that samples with low PD exhibit lower perplexity. This observation aligns with the trends illustrated in Figure \ref{3B_train_loss}, where training with low-PD data leads to a rapid decrease in training loss during the initial stages, providing the model with a clear direction for gradient optimization. Notably, even when handling with high-PD data in later stages, the model maintains steady loss reduction, ultimately achieving a minimized training loss.

\subsection{Spearman correlation coefficient of PDs from different RM sizes} \label{c3}
\begin{figure}[htbp]
    \centering
\includegraphics[width=\linewidth]{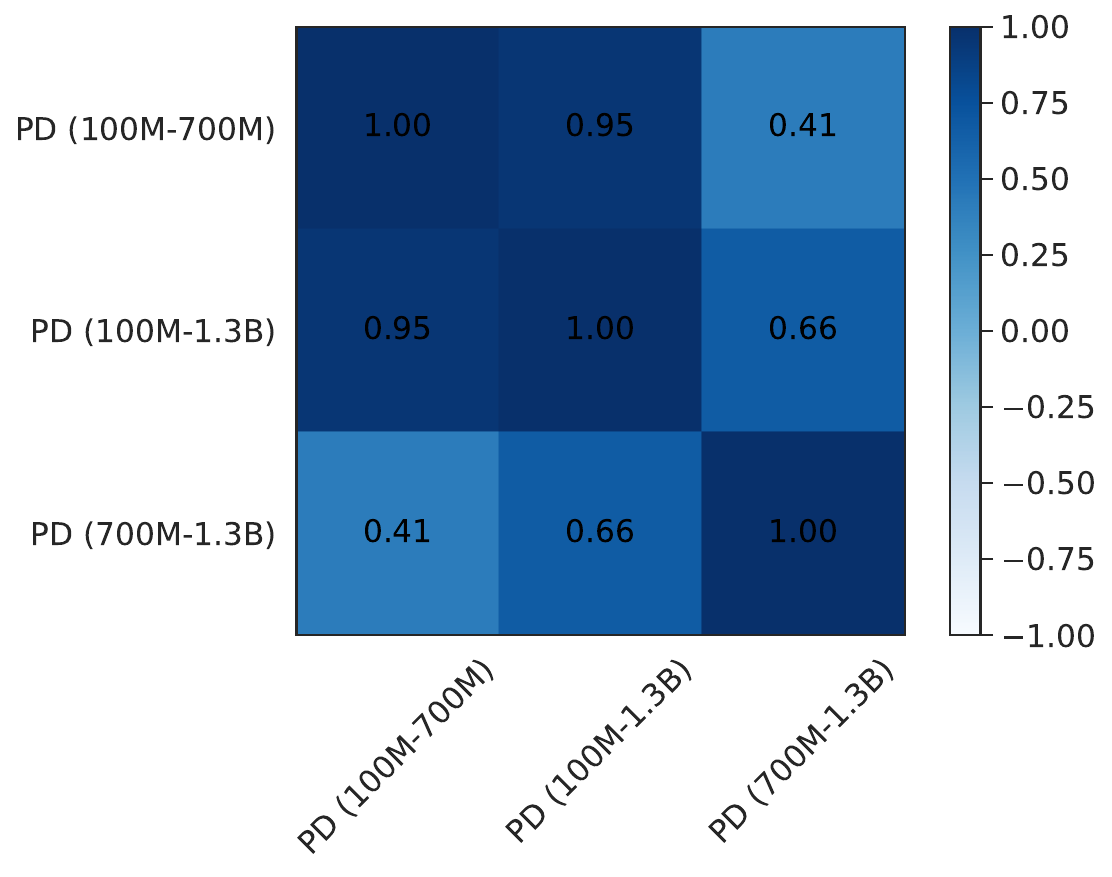}
    \caption{Spearman correlation coefficient of PDs from different RM sizes.}
    \label{corr_heat_map}
\end{figure}

We evaluate the Spearman correlation coefficients between different PD types, as shown in Figure \ref{corr_heat_map}, and find a strong correlation among PDs derived from RMs of varying sizes, which indicates calculating PD with smaller RMs produces results consistent with larger RMs, optimizing computational resources.

\subsection{PD distribution across different domains}
\label{c2}
\begin{figure*}[htbp]
    \centering
    \includegraphics[width=0.8\linewidth]{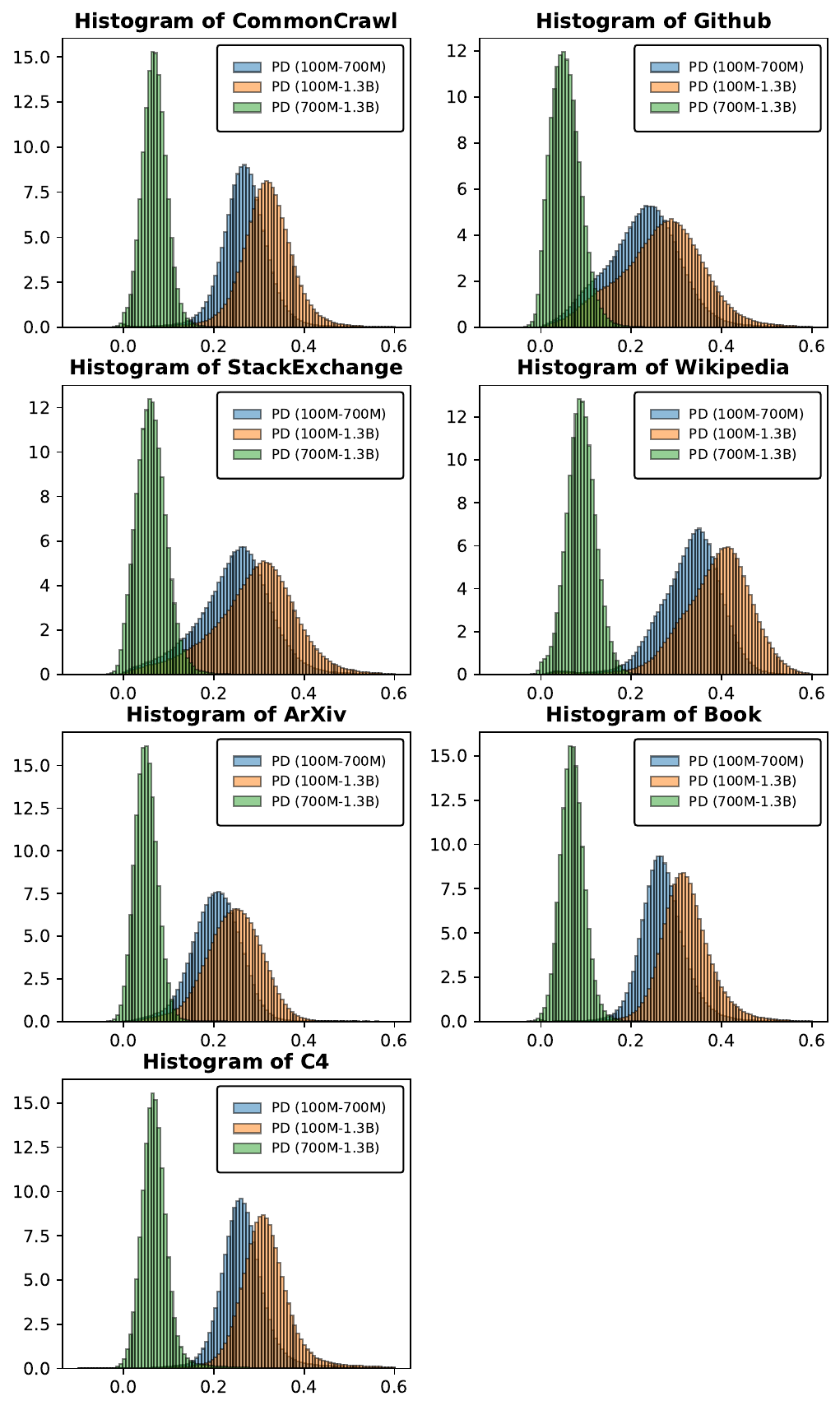}
    \caption{PD distribution across different domains.}
    \label{PD__distribution}
\end{figure*}

As illustrated in Figure \ref{PD__distribution}, large size discrepancies among RMs result in broader PD distributions, which enhance data differentiation. This finding is supported by ablation tests, where the 100M-1.3B PD calculations yielded the best results. Additionally, PD maintains a consistent distribution across domains. This stability makes PD a reliable metric. Organizing training data by PD ensures it does not skew towards specific sub-domains, allowing the model to encounter diverse data at every stage.

\subsection{Detailed performance on the benchmarks}
\begin{figure*}[ht]
    \centering
    \begin{subfigure}{0.49\linewidth}
        \centering
        \includegraphics[width=\linewidth]{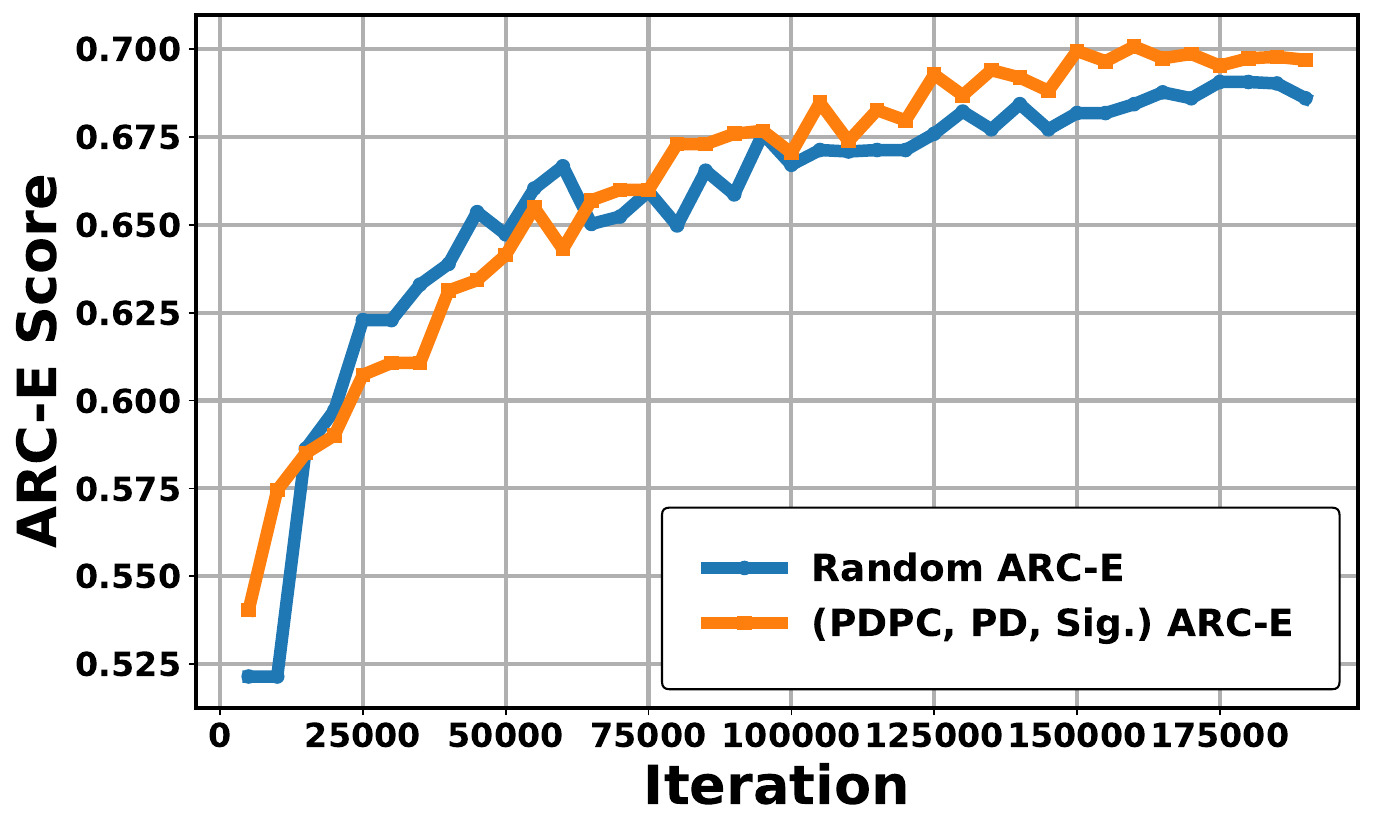}
        \caption{Accuracy on ARC-E.}
        \label{fig:arc-e}
    \end{subfigure}
    \hfill
    \begin{subfigure}{0.49\linewidth}
        \centering
        \includegraphics[width=\linewidth]{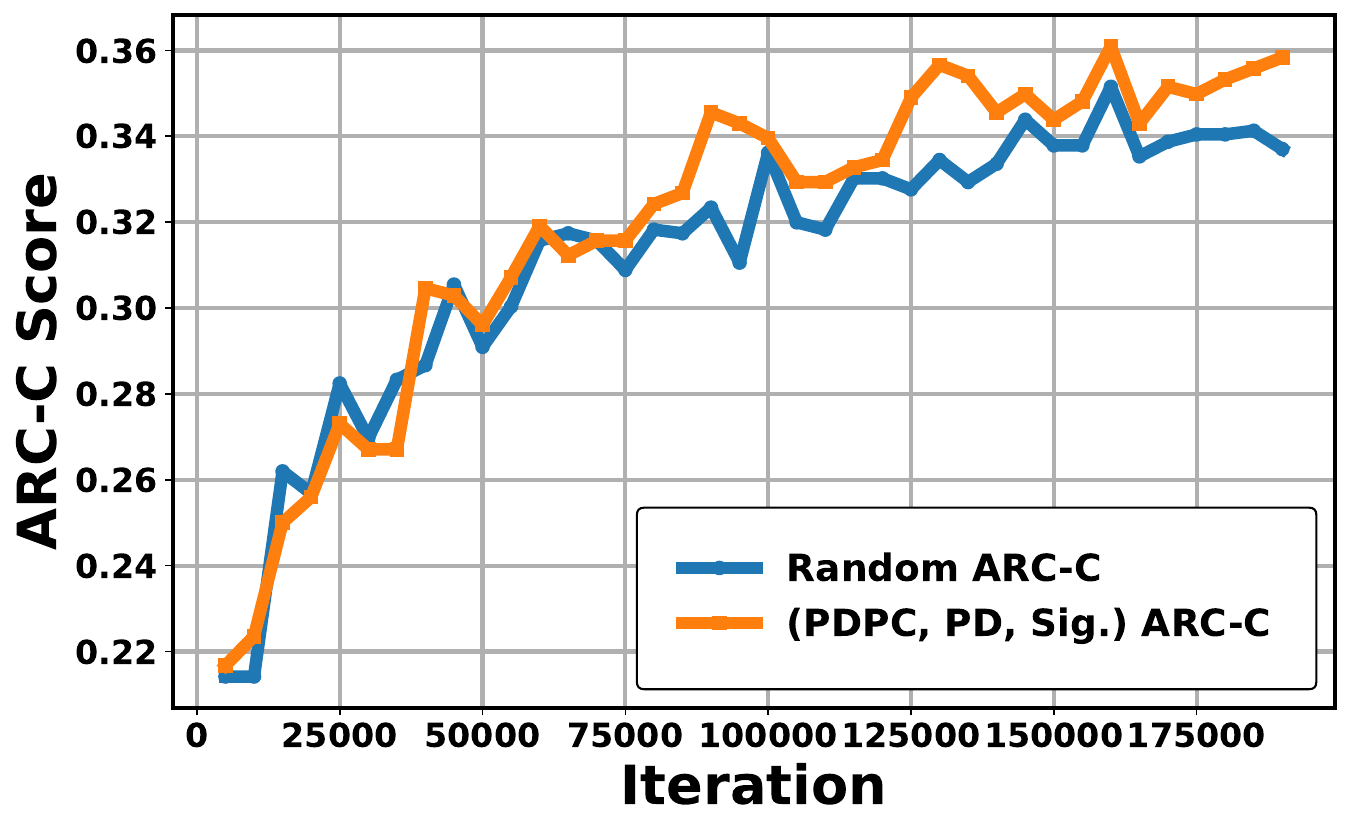}
        \caption{Accuracy on ARC-C.}
        \label{fig:arc-c}
    \end{subfigure}
    \hfill
    \begin{subfigure}{0.49\linewidth}
        \centering
        \includegraphics[width=\linewidth]{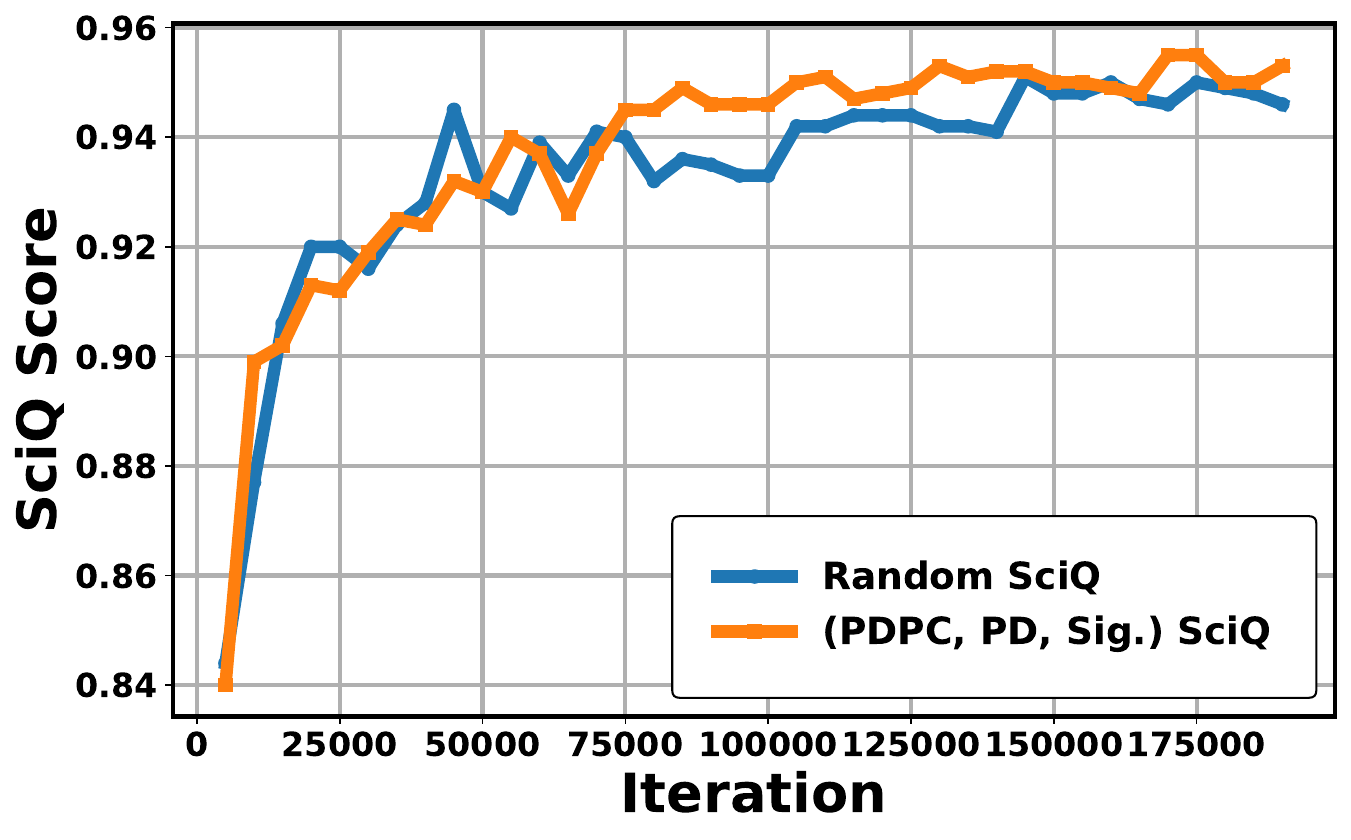}
        \caption{Accuracy on SciQ.}
        \label{fig:sciq}
    \end{subfigure}
    \begin{subfigure}{0.49\linewidth}
        \centering
        \includegraphics[width=\linewidth]{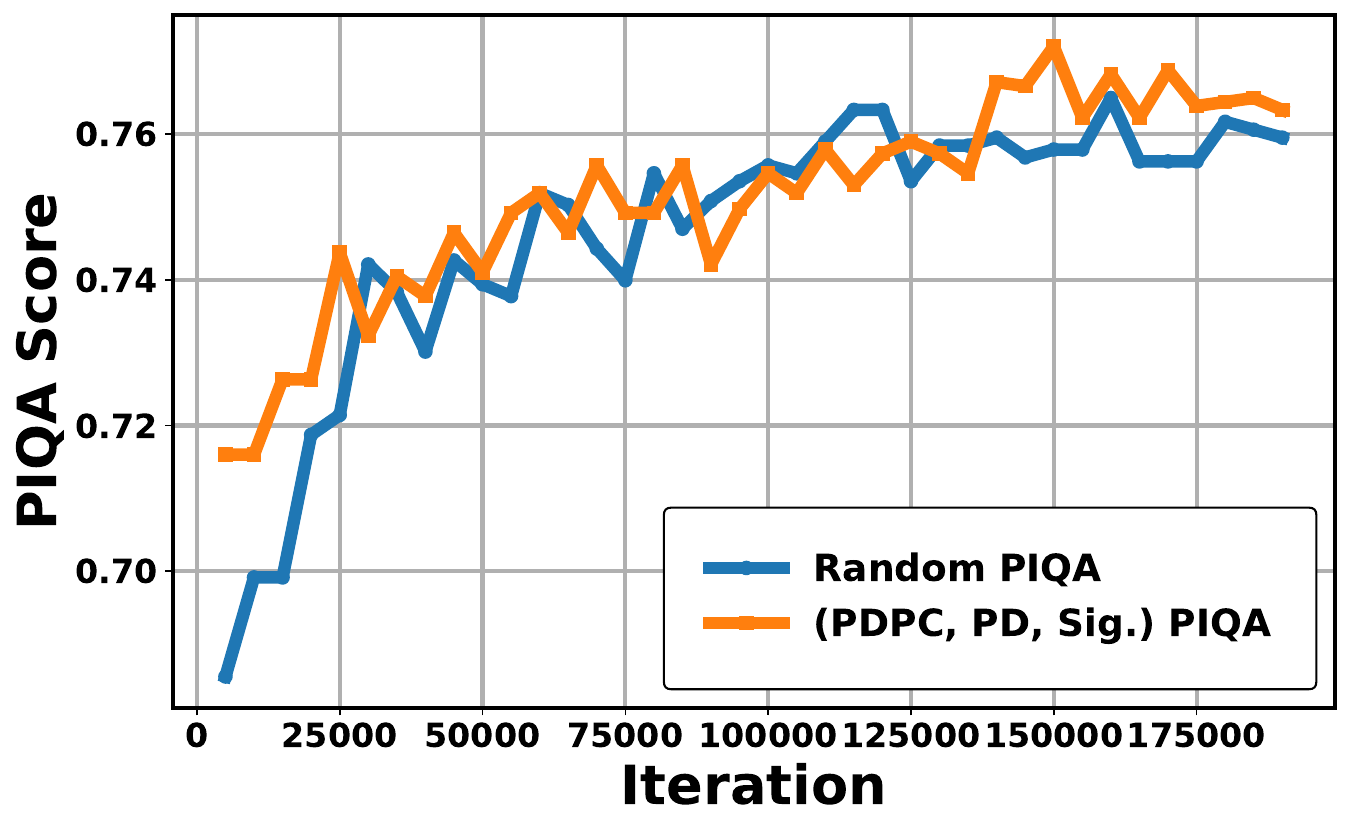}
        \caption{Accuracy on PIQA.}
        \label{fig:piqa}
    \end{subfigure}
    \begin{subfigure}{0.49\linewidth}
        \centering
        \includegraphics[width=\linewidth]{figure/MMLU.pdf}
        \caption{Accuracy on MMLU.}
        \label{fig:mmlu}
    \end{subfigure}
    \begin{subfigure}{0.49\linewidth}
        \centering
        \includegraphics[width=\linewidth]{figure/CMMLU.pdf}
        \caption{Accuracy on CMMLU.}
        \label{fig:cmmlu}
    \end{subfigure}
    \begin{subfigure}{0.49\linewidth}
        \centering
        \includegraphics[width=\linewidth]{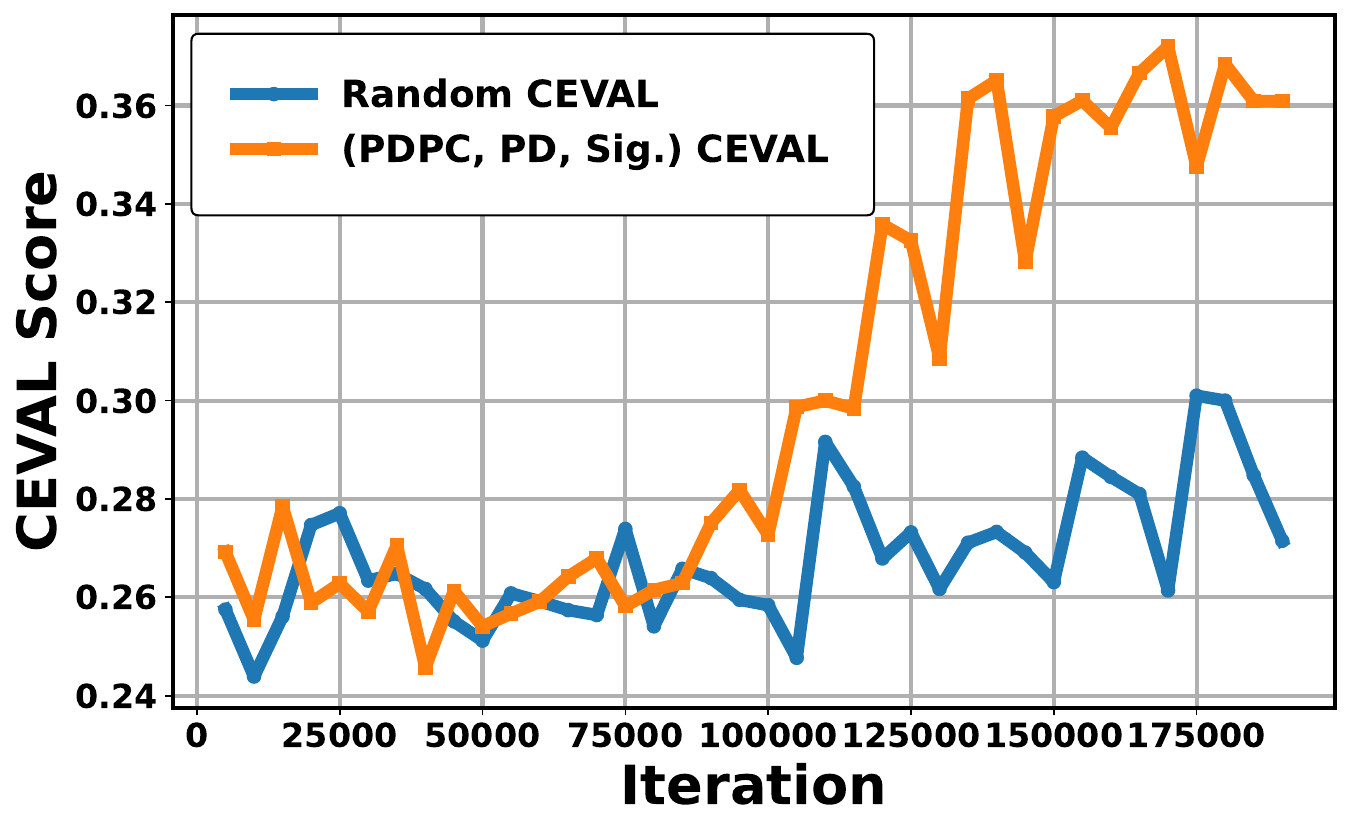}
        \caption{Accuracy on CEVAL.}
        \label{fig:ceval}
    \end{subfigure}
    \caption{Few-shot downstream performance on various benchmarks with respect to pretraining iterations for Random and $\textit{PDPC-PD-S.}$. We train a 3B model over 1T tokens, demonstrating superior performance with our approach.} 
    \label{fig:overall}
\end{figure*}

In this section, we explore the detailed performance metrics across various benchmarks. Figure \ref{fig:overall} illustrates how these metrics evolve during training, focusing specifically on few-shot downstream performance. We compare the performance of \textit{Random} and $\textit{PDPC-PD-S.}$, across pretraining iterations.

Our experiments involved training a model with 3 billion parameters on a dataset containing 1 trillion tokens. This large-scale training setup effectively demonstrates our approach's superior performance. The results highlight the gains in accuracy and performance achieved by our method, showcasing its clear advantages over \textit{Random}. We affirm the potential of the $\textit{PDPC-PD-S.}$ methodology in enhancing model performance, particularly when faced with diverse and challenging benchmarks.
\onecolumn
\subsection{Prompts for Case Study}
\label{sec:prompt}
The prompt and specific rules used in Section \ref{sec:case} to analyze the linguistic features of data across different PD intervals are as follows.
\begin{tcolorbox}[colback=white!95!gray,colframe=gray!50!black,rounded corners,label={prompt-dot}, title={Prompts for Property Recognition}]
\begin{lstlisting}[breaklines=true, xleftmargin=0pt, breakindent=0pt, columns=fullflexible, mathescape, numbers=none]
You are a language model training data annotator. Your task is to identify whether the given text possesses the following characteristic: {Property}

The text to be annotated is:
{text}

Please determine whether the given text possesses this characteristic according to the above rules. The output format should be "Because..., my answer is 'X'." where X must be either "yes" or "no." You should remain objective and refrain from adding any further comments after making your choice.
\end{lstlisting}
\end{tcolorbox}
\{Property\} is from one of the following rules:
\begin{tcolorbox}[colback=white!95!gray,colframe=gray!50!black,rounded corners,label={prompt-dot2}, title={Rules for Property Recognition}]
\begin{lstlisting}[breaklines=true, xleftmargin=0pt, breakindent=0pt, columns=fullflexible, mathescape, numbers=none]
1. Does the text contain polysemous words? Polysemous words may make understanding more difficult.

2. Does the text use specialized terminology? Specialized terminology may require specific domain knowledge to understand.

3. Does understanding the text require specific cultural background knowledge? Cultural background dependence may increase the complexity of understanding.

4. Does the text require logical reasoning to understand? Logical reasoning adds depth to understanding.

5. Does the text contain elements of humor? Humor may affect the way the text is understood.

6. Does the text explore ethical or moral issues? This may increase the depth of thought.

7. Does the text use complex sentence structures? Complex sentence structures may increase the difficulty of understanding.

8. Does the text contain scientific or technical concepts? These concepts may require specific knowledge to understand.

9. Does the text express obvious emotional tones? Emotional tones may affect the understanding of the text.

10. Does understanding the text require additional background knowledge? Background knowledge requirements may affect the comprehensibility of the text.
\end{lstlisting}
\end{tcolorbox}
\twocolumn

\onecolumn
\section{Case Study}

\begin{longtable}{p{15cm}}
\caption{Samples are divided into 10 PD quantiles, with two samples representing each quantile.}
\label{PD_quantiles} \\
\toprule
\midrule
\endfirsthead

\caption[]{(continued)} \\
\toprule
\midrule
\endhead

\midrule
\endfoot

\bottomrule
\endlastfoot

\small \textbf{0-10\%} \\
\midrule
\small \textit{\textbf{Sample 1}:} The need to practice good self-care doesn't change in this working environment, but how you accomplish this goal might. Much of Arel's own self-care regimen needed to be adjusted."I was used to weekly massage and monthly chiropractic care. That was gone," she explains. "I am used to runs and yoga and time to meditate in complete silence. That was gone, too." ... \\
\small \textit{\textbf{Sample 2}:} I have to ask you, why'd you--wha--wha--why are you peeing right here?Creepy Guy: What?Kumar: I mean... why'd you pee right next to me when you could like, choose that bush, or--?Creepy Guy: Well, this bush looked like I should pee on it. Why are you peeing on it?Kumar: Well, no one was here when I chose this bush.Creepy Guy: Oh, so you get to pee on it and no one else does? Huh?Kumar: ... \\
\midrule
\small \textbf{10-20\%} \\
\midrule
\small \textit{\textbf{Sample 1}:} boolean insertventas() String sql "INSERT INTO ventas (id\_venta, venFechaventa, venId\_cliente, venIdadministrador, venTotalventa) VALUES (NULL, '" + vent() "', '" clasu.getId\_usuarios() "', '1', '" + pnlProductos.totall + "')"; try con cn.getConnection(); ps = con.prepareStatement(sql); ps.executeUpdate(); return true; catch (SQLException ex) Logger.getLogger(LogicaSql.class... \\
\small \textit{\textbf{Sample 2}:} In terms of providing shorter stay parking, Bell Street multi storey car park is identified as a long stay car park, and the tariffs are so designed to encourage the use of the facility by all day / half day parkers with more flexible tariffs available at other car parks and the on street spaces around the vicinity allow for parking for up to one hour.I have commented that there is no short term (30 minutes to 2 hour) parking available at the West Bell Street multi-storey car park and ... \\
\midrule
\small \textbf{20-30\%} \\
\midrule
\small \textit{\textbf{Sample 1}:} Maybe it just sagslike a heavy load. Or does it explode? ~ by Langston HughesIn 1849, Elizabeth Blackwell became the first woman to graduate from a U.S. medical school in N.Y.In 1864, Rebecca Lee Crumpler became the first black woman to graduate from a U.S. medical school in New England.In 1915, women represented approximately 5\% of the physician workforce in the U.S.In 1983, women represented approximately 1/3 of U.S. medical school matriculants ... \\
\small \textit{\textbf{Sample 2}:} FILED UNDER SEAL PURSUANT TO PROTECTIVE ORDER rise to a direct infringement claim against it. See, e.g., Akami Techs., Inc. v. Limelight Networks, Inc., 797 F.3d 1020, 1023 (Fed. Cir. 2015) (noting entities are liable for performance they control). The evidence further shows that Badoo Software Limited and Badoo Limited are also intimately involved in Badoo Trading's creation and ownership of the infringing Bumble application... \\
\midrule
\small \textbf{30-40\%} \\
\midrule
\small \textit{\textbf{Sample 1}:} ... I myself should be a castaway.Young's Literal: but I chastise my body, and bring it into servitude, lest by any means, having preached to others -- I myself may become disapproved.As noted earlier, Paul now applies the example from the Greek sports arena directly to himself ("I discipline… I myself") and does so that he might present himself as an example or model for other believers to imitate (cp 1Co 4:16, 11:1, 1Th 1:6, cp Heb 6:12, He 13:7, 3Jn 1:11)...  \\
\small \textit{\textbf{Sample 2}:}  ... (B) of from about 0.1 to about 10.0\% w/v of a bioadhesive polymeric stabilizer selected from the group consisting of:(i) polyethylene-polypropylene oxide tri-block co-polymers of the formula;(polyethylene oxide)a -(polypropylene oxide)b -(polyethylene oxide)c wherein PA4 a is 46, 52, 62, 75, 97, 98, 122, or 128; PA4 b is 16, 30, 35, 39, 47, 54, or 67; and PA4 c is 46, 52, 62, 75, 97, 98, 122, or 128;(ii) polyvinyl alcohol,(iii) polyvinyl pyrr......\\
\midrule
\small \textbf{40-50\%} \\
\midrule
\small \textit{\textbf{Sample 1}:}  ... as foreigners seem to have trouble believing about the trees. A second year passed before the fruit split open, and I came out, and several siblings, with hair like Sapham and wings like Pham, and we have no gender because we are not animals but fruit and we like to sing, too, and we like to fly, and we like to be loyal, and we like to love. The tree opened up and flew away and when it was done only twigs and a few blue leaves remained, and then they blew away, too, and we were all born, and ready to live... \\
\small \textit{\textbf{Sample 2}:}  ... She had a World Series poker face, and it never slipped. He wondered if she'd had a plan of her own, given how long she went without looking rattled. Maybe she assumed he was in Lenny's corner. Maybe she'd been lining up a double hit.Looking at it now though, it surprised him, how easily he'd committed to killing Lenny. Not that he regretted it, but clearly he was risking fatal penalties, stepping in on a Garcia job and smoking one of their guys. He could live with the risk, but he'd never thought about it at the time...\\
\midrule
\small \textbf{50-60\%} \\
\midrule
\small \textit{\textbf{Sample 1}:}   "We had him just where we wanted him—but it's a fine line." Were they talking about him? They must be. Or was it just arrogance to think that? "I really don't see what has changed," Cheng Li said. "If anything, we're closer to the result we want." Connor felt his head begin to pound. If they were talking about him, what did this mean? Had they had something to do with what had happened to Grace? He remembered in a flash Grace saying that Cheng Li had known her plan...  \\
\small \textit{\textbf{Sample 2}:}  ...in which, I suspect, the most diverse directions of my work will come together.Added to this is an external incentive. Next year Toledo, I hear, is to be the scene of a big Greco exposition: not only would I like scrupulously to avoid this occasion, I fear that this hitherto still so uninterrupted earthly constellation, which is Toledo, will after this congestion be left changed, popularized, so that this is almost the last moment for surprising it in its remoteness.Now it goes against me, dear friend, to give in to this important decision...\\
\midrule
\small \textbf{60-70\%} \\
\midrule
\small \textit{\textbf{Sample 1}:} ... The stringy, yet short, dark-skinned Mawikizi returned the salute with a smile. "I pulled some serious strings to haul you out here this quick, Keyes." He held the door open for Keyes, and it banged shut behind them once the lieutenant stepped through. "Walk with me." The rough rock-tunneled corridor stretched out in front of them. Mawikizi led Keyes down past offices, shouldering past privates and officers who stood to attention as he walked by. Keyes glanced off down a subcorridor, seeing barracks in the distance...  \\
\small \textit{\textbf{Sample 2}:}  ... The United States District Court (federal) hires court reporters for its courts, including those within New York State. No test is required. When vacancies occur, announcements for experienced reporters are posted in places such as the NYSCRA website. Reporters who meet stated criteria are told how to apply. Selected candidates are rigorously interviewed for appointment to this important judicial arena. Realtime certification has become a prerequisite for most federal court reporting positions...\\
\midrule
\small \textbf{70-80\%} \\
\midrule
\small \textit{\textbf{Sample 1}:}  ... For example, though the Gaon railed against the potentially negative effects of synagogue attendance, it is obvious that women did go to the synagogue. Few shared his jaundiced view, though others did point out the possible pitfalls. Similarly, the Gaon's horror at the prospect of his daughters strolling in the street could not be a guide for the many women who spent their days pursuing their family's livelihood in the marketplace.Beyond the sphere of ritual behavior, a woman was expected to fulfill a religious role analogous to her social function and reflecting her status in society—that is, woman as religious facilitator...  \\
\small \textit{\textbf{Sample 2}:}  ... Assuming that the amplitude is larger than $b_{AC}^{th}$, the switching time is determined by the frequency sweeping rate $\alpha$. Once the magnetic moment is captured intoautoresonance, its nonlinear precession frequency is locked to the instantaneous excitation frequency $f(t)=f_0+\alpha t$ (remember that $\alpha<0$).If we define the switching time $\tau$ as the time it takes for the moment to cross the energy barrier and knowing that the frequency vanishes atthe top of the barrier. At the top of the energy barrier, the precession reverses from counter-clockwise toclockwise...\\
\midrule
\small \textbf{80-90\%} \\
\midrule
\small \textit{\textbf{Sample 1}:} ... The manipulation and processing of stereo image sequences demand higher costs in memory storage, transmission bandwidth, and computational complexity than of monoscopic images. This chapter investigates scenarios for cost reduction by using reversible watermarking. The basic principle is to embed some data by reversible watermarking instead of either computing or storing/transmitting it. Storage and/or bandwidth are reduced by embedding into one frame of a stereo pair the information needed... \\
\small \textit{\textbf{Sample 2}:}  ...Rosengarten pitched the third and fourth innings and Guillozet closed it out. Barhorst finished 2 for 2 at the plate. The Tigers tied in a nonconference game on Monday in Covington. Jackson Center scored two runs in the top of the seventh but Covington scored four runs in the bottom of the inning to tie it. The Tigers scored two runs in the third in three in the fourth, but the Buccaneers scored five in the fifth to tie it 5-5. Jackson Center took the lead with two runs in the sixth. Jackson Center had 10 hits and five errors while Covington had three hits and three errors...\\
\midrule
\small \textbf{90-100\%} \\
\midrule
\small \textit{\textbf{Sample 1}:}  ... In fact, it might be fair to say that in this member of an inferior race, there were as many animal characteristics as human ones, but they were gentle and caressing animal ways. He had nothing of the wild animal in him, but rather the physiognomy of a good and faithful dog, like a courageous Newfoundland dog, who can become man's friend and not just his companion. Indeed, he came at the sound of his name, like one of those devoted animals, to rub himself against the master whose hand gripped his own... \\
\small \textit{\textbf{Sample 2}:} ...the whole world has transformed right into a Global City. Details is passed into every corner of the world within minutes. This raising appeal gave rise to numerous information as well as material organizing websites on the net .Wpengine deals pay as you go August 2018 Web holding service is a solution which enables the companies as well as individuals to put information and web content online. . It has many kinds and also groups. Following are its main categories. Wpengine deals pay as you go August 2018 Exactly what is the objective of web organizing...\\
\end{longtable}

\end{document}